\pdfoutput=1

\documentclass[10pt, journal,twoside]{IEEEtran}

\ifCLASSINFOpdf
\else
\fi

\usepackage{cite}
\usepackage{amsmath}
\usepackage{amsfonts}
\usepackage{threeparttable}
\usepackage{graphicx}
\usepackage{subfigure}
\usepackage{color}
\usepackage{float}
\usepackage{booktabs}
\usepackage{makecell}
\usepackage{multirow}
\usepackage{pifont}
\usepackage{amssymb}
\usepackage{mathrsfs}

\usepackage[ruled, linesnumbered]{algorithm2e}

\begin{document}

%
% paper title
% Titles are generally capitalized except for words such as a, an, and, as,
% at, but, by, for, in, nor, of, on, or, the, to and up, which are usually
% not capitalized unless they are the first or last word of the title.
% Linebreaks \\ can be used within to get better formatting as desired.
% Do not put math or special symbols in the title.
\title{HKNAS: Classification of Hyperspectral Imagery Based on Hyper Kernel Neural Architecture Search}
%
%
% author names and IEEE memberships
% note positions of commas and nonbreaking spaces ( ~ ) LaTeX will not break
% a structure at a ~ so this keeps an author's name from being broken across
% two lines.
% use \thanks{} to gain access to the first footnote area
% a separate \thanks must be used for each paragraph as LaTeX2e's \thanks
% was not built to handle multiple paragraphs
%

\author{Di Wang,~\IEEEmembership{Student Member,~IEEE,}
        Bo Du,~\IEEEmembership{Senior Member,~IEEE,}
        Liangpei Zhang,~\IEEEmembership{Fellow,~IEEE,}\\
        and Dacheng Tao,~\IEEEmembership{Fellow,~IEEE}% <-this % stops a space
\thanks{\textit{Corresponding authors: Bo Du and Liangpei Zhang.}}
\thanks{D. Wang and B. Du are with the School of Computer Science, Wuhan University,
        Wuhan 430072, China (e-mail: wd74108520@gmail.com; gunspace@163.com).}
\thanks{L. Zhang is with the State Key Laboratory of Information
Engineering in Surveying, Mapping and Remote Sensing, Wuhan University,
Wuhan 430079, China (e-mail: zlp62@whu.edu.cn).}
\thanks{D. Tao is with the JD Explore Academy, China and is also with the School of Computer Science, Faculty of Engineering, The University of Sydney, Australia (dacheng.tao@gmail.com).}% <-this % stops a space
% <-this % stops a space
%\thanks{Manuscript received April 19, 2005; revised August 26, 2015.}
}

% note the % following the last \IEEEmembership and also \thanks - 
% these prevent an unwanted space from occurring between the last author name
% and the end of the author line. i.e., if you had this:
% 
% \author{....lastname \thanks{...} \thanks{...} }
%                     ^------------^------------^----Do not want these spaces!
%
% a space would be appended to the last name and could cause every name on that
% line to be shifted left slightly. This is one of those "LaTeX things". For
% instance, "\textbf{A} \textbf{B}" will typeset as "A B" not "AB". To get
% "AB" then you have to do: "\textbf{A}\textbf{B}"
% \thanks is no different in this regard, so shield the last } of each \thanks
% that ends a line with a % and do not let a space in before the next \thanks.
% Spaces after \IEEEmembership other than the last one are OK (and needed) as
% you are supposed to have spaces between the names. For what it is worth,
% this is a minor point as most people would not even notice if the said evil
% space somehow managed to creep in.

% The paper headers
\markboth{Journal of \LaTeX\ Class Files,~Vol.~14, No.~8, August~2015}{Wang \MakeLowercase{\textit{et al.}}: HKNAS FOR HSI CLASSIFICATION}
% The only time the second header will appear is for the odd numbered pages
% after the title page when using the twoside option.
% 
% *** Note that you probably will NOT want to include the author's ***
% *** name in the headers of peer review papers.                   ***
% You can use \ifCLASSOPTIONpeerreview for conditional compilation here if
% you desire.

% If you want to put a publisher's ID mark on the page you can do it like
% this:
%\IEEEpubid{0000--0000/00\$00.00~\copyright~2015 IEEE}
% Remember, if you use this you must call \IEEEpubidadjcol in the second
% column for its text to clear the IEEEpubid mark.

% use for special paper notices
%\IEEEspecialpapernotice{(Invited Paper)}

% make the title area
\maketitle

% As a general rule, do not put math, special symbols or citations
% in the abstract or keywords.
\begin{abstract}

  Recent neural architecture search (NAS) based approaches have made great progress in hyperspectral image (HSI) classification tasks. However, the architectures are usually optimized independently of the network weights, increasing searching time and restricting model performances. To tackle these issues, in this paper, different from previous methods that extra define structural parameters, we propose to directly generate structural parameters by utilizing the specifically designed hyper kernels, ingeniously converting the original complex dual optimization problem into easily implemented one-tier optimizations, and greatly shrinking searching costs. Then, we develop a hierarchical multi-module search space whose candidate operations only contain convolutions, and these operations can be integrated into unified kernels. Using the above searching strategy and searching space, we obtain three kinds of networks to separately conduct pixel-level or image-level classifications with 1-D or 3-D convolutions. In addition, by combining the proposed hyper kernel searching scheme with the 3-D convolution decomposition mechanism, we obtain diverse architectures to simulate 3-D convolutions, greatly improving network flexibilities. A series of quantitative and qualitative experiments on six public datasets demonstrate that the proposed methods achieve state-of-the-art results compared with other advanced NAS-based HSI classification approaches.

\end{abstract}

% Note that keywords are not normally used for peerreview papers.

% For peer review papers, you can put extra information on the cover
% page as needed:
% \ifCLASSOPTIONpeerreview
% \begin{center} \bfseries EDICS Category: 3-BBND \end{center}
% \fi
%
% For peerreview papers, this IEEEtran command inserts a page break and
% creates the second title. It will be ignored for other modes.
\IEEEpeerreviewmaketitle

\section{Introduction}
% The very first letter is a 2 line initial drop letter followed
% by the rest of the first word in caps.
% 
% form to use if the first word consists of a single letter:
% \IEEEPARstart{A}{demo} file is ....
% 
% form to use if you need the single drop letter followed by
% normal text (unknown if ever used by the IEEE):
% \IEEEPARstart{A}{}demo file is ....
% 
% Some journals put the first two words in caps:
% \IEEEPARstart{T}{his demo} file is ....
% 
% Here we have the typical use of a "T" for an initial drop letter
% and "HIS" in caps to complete the first word.
\IEEEPARstart{T}{h}e hyperspectral image (HSI) possesses abundant spectral and spatial information, which are separately characterized by hundreds of continuous and narrow bands in large wavelength ranges and high spatial resolution pixels that provide clear ground landscape descriptions. These characteristics facilitate the understanding and interpretation of the observed whole scene. Thus, the HSIs have been widely applied in many fields, such as precision agriculture \cite{agriculture_1} and environmental monitoring \cite{env_monit_1}. Among these tasks,  recognizing specific and unique semantic categories for every pixel is one of the most fundamental purposes, as is called HSI classification.

Many early methods, which are simple classifiers such as k-nearest neighbors \cite{KNNinHSIclassify}, multinomial logistic regression \cite{MLR}, and support vector machine (SVM) \cite{LRinHSIclassify}, have been used to directly reach the category of target pixels only using the spectral feature by directly regarding the values of all bands as compositions of a spectral vector. However, the performances of these methods are unsatisfactory because of the lacking of neighborhood information related to surrounding objects. To tackle this issue and further improve the classification, spatial properties are introduced into many subsequent approaches \cite{MP_1,Fang2015,Gu2012}, where the SVM is the most commonly used classifier in the above methods since it performs stable even if processing high-dimensional data such as the HSI. Therefore, in our later performance comparison studies, the SVM is evaluated as the baseline to represent conventional methods. 

The above methods all can be seen as the pattern of feature engineering that generates discriminative features through the spectral-spatial information lying in HSI for effective classification. However, according to \cite{representlearning}, these methods actually belong to the shallow-layer model, causing the generated features to be considered as shallow features, which are not robust in complex circumstances and not able to describe essential characteristics well. Additionally, these methods need to be manually developed with the experience of the designer. Thus, the performances of the obtained handcrafted features strongly depend on artificial factors.

To tackle these problems, with excellent characteristics that automatically extract deep features reflecting the inherent properties of target objects, many deep learning technologies gradually dominate the HSI classification field. Among these technologies, with the distinguished characteristics of local perception and weight sharing, the convolutional neural network (CNN) becomes the most representative and widely used framework \cite{hu2015deep, 3dcnn, assmn}. Besides these pixel-level classification methods, CNNs have also been extensively applied to image-level HSI classification \cite{ssfcn, freenet, fcontnet} through a family of special networks --- fully convolutional networks (FCNs) that only include convolutional or pooling layers. Compared with pixel-level classification methods that receive spectral vectors or spatial patches of the target pixel and produce corresponding categories, the image-level HSI classification usually receives whole images and outputs corresponding classification maps, meaning all pixels are simultaneously classified. This is similar to the semantic segmentation task in the computer vision field. In this paper, when constructing the image-level network, we follow the configurations of high-resolution retaining FCN, which is similar to the backbone network in \cite{danet}. The relevant details will be introduced later.

Although current deep learning-based methods perform well, they are still not intelligent enough since the weights are trained under the premise of given architectures. For example, some classical architectures, such as the inception layer of GoogLeNet \cite{inception_v1} and the residual block in ResNet \cite{resnet} are both constructed with the help of expert experiences. Designing such a module usually requires long-time explorations and inevitably introduces man-made biases. So can we further break up the limitations of artificial constructions and let the machine automatically find the most suitable network structure?

The neural architecture search (NAS) technology aims to automate architecture engineering procedures, where the whole network including weights and structures can be naturally built without any manual interventions after defining search spaces and search algorithms. Early NAS methods are unfriendly for individual researchers since the adopted discrete search strategies such as reinforcement learning (RL) and evolutionary algorithm (EA) force all architectures need to be independently sampled, trained, and evaluated, bringing unaffordable searching costs. Most of the currently popular NAS methods are one-shot NAS \cite{darts, pdarts, pcdarts}, which involves a hyper network that contains all candidate architectures. Since architecture variables are represented by relaxed structural parameters, which can be optimized by continuous optimization strategies such as the gradient algorithm. Architecture searching procedures can be regarded as constant adjustments of structural parameters, and the obtained architectures are derived from final optimization results. One-shot means only the final generated architecture needed to be evaluated, greatly reducing the search time. Up to now, some one-shot NAS-based approaches have been proposed for HSI classification \cite{autocnn_hsi,3danas,sstn_nas}.

\begin{figure}[t]
  \centering
  \subfigure[]{\includegraphics[width=0.7\linewidth]{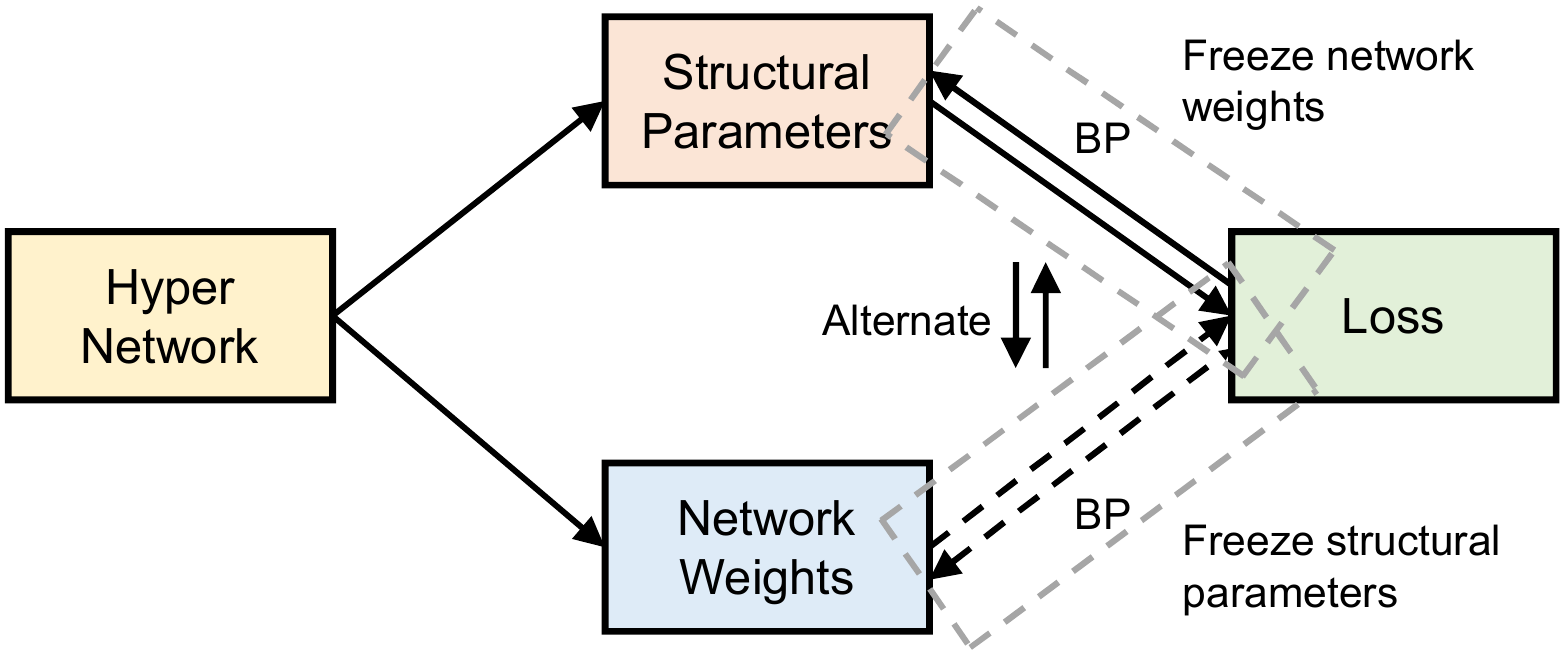}}\\
  \subfigure[]{\includegraphics[width=0.7\linewidth]{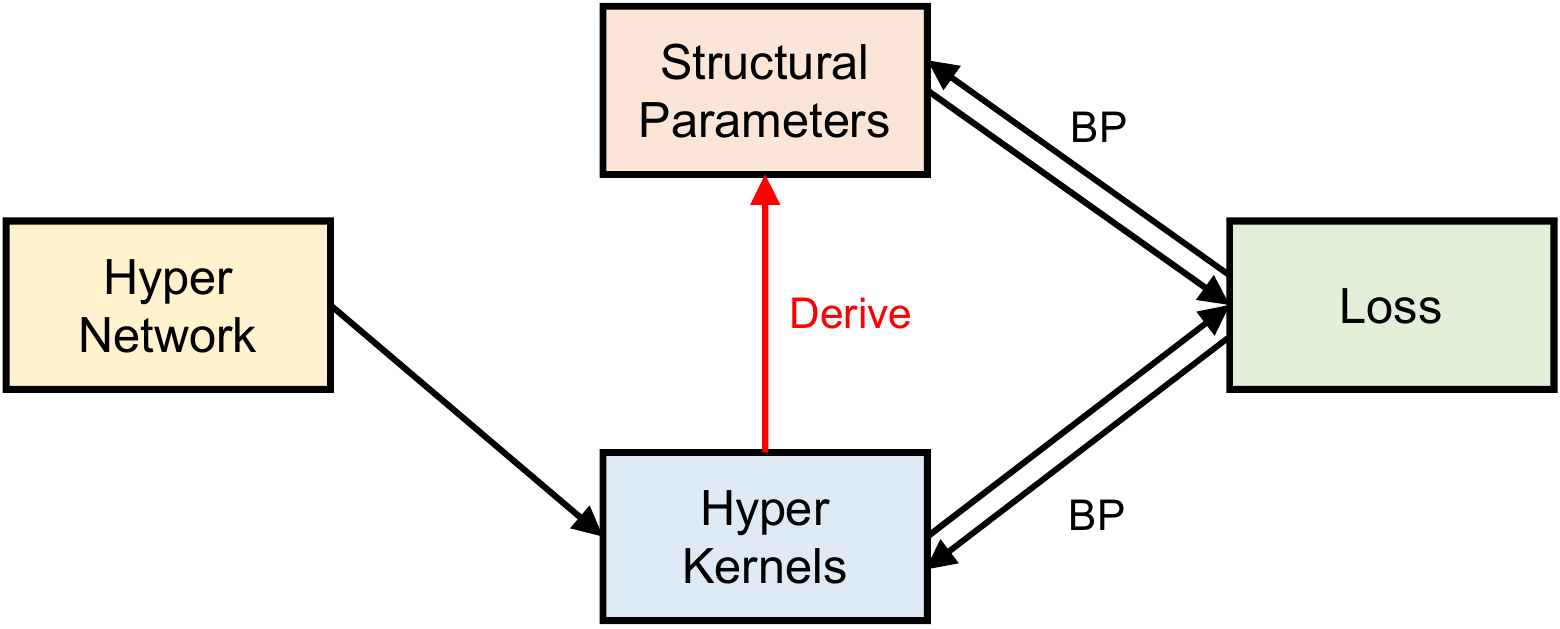}}
  \caption{The framework of the searching stage in different one-shot NAS methods. (a) Previous one-shot NAS represented by DARTS. (b) Proposed HKNAS. Previous one-shot NAS separately defines structural parameters and weights in the hyper network, where these two parts are alternately updated, formatting a complex dual optimization problem. Here, we show a situation of adjusting structural parameters by freezing network weights. While in HKNAS, structural parameters are generated by the predefined hyper kernels instead of independently claimed. Thus, the whole framework can be updated as a simple one-tier optimization problem. BP means back propagation.}
  \label{oneshotnas}
\end{figure}

However, the above one-shot NAS methods still face problems. Recalling the searching procedure of a representative one-shot NAS solution---DARTS \cite{darts}: structural parameters and network weights are independently defined and separately optimized on the split training and validation sets by regarding the architecture searching task as a complex dual optimization problem. Specifically, structural parameters are frozen when updating network weights, and vice versa. Intuitively, network architectures and weights affect each other. It is no doubt that after determining the final architecture, the obtained network needs to be trained from scratch before evaluation. Thus, the core purpose of the search stage is to find a suitable architecture, and this issue is transformed into how to optimize structural parameters. What we doubt is can these independently defined structural parameters really represent the architecture? In our consideration, the effectiveness of an architecture strongly relates to the distinguishability of features generated by selected operations, while the features are associated with operators, demonstrating that architectures should be relevant to the operators. But the independently defined structural parameters ignore these relationships, or the structural parameters only relate to the network weights through the weak connections of loss computation and back propagation, since freezing one part cannot hinder the update of another part. In addition, each edge in the hyper network contains all candidate operations, meaning all operators such as convolutional kernels need to be simultaneously defined and updated when training the hyper network, requiring too many trainable parameters, and the method practicability is limited.

Regarding the above issues, we propose a new one-shot NAS method for HSI classification. In order to highlight the relationships between architectures and network weights. We directly derive structural parameters from network weights (also called hyper kernels, which will be introduced later). Concretely, different areas of one kernel can generate structural parameters corresponding to different architectures since the weights in different areas aggregate the local contexts of different locations. Then, architectures are determined based on the optimized structural parameters, converting the complex dual optimization problem to a one-tier optimization problem that only needs to optimize kernel weights, decreasing the searching difficulty and is convenient for obtaining better architectures. The differences between our NAS method and previous one-shot NAS approaches in the design of structural parameters are shown in Figure \ref{oneshotnas}. Moreover, following \cite{singlepathnas}, different from previous methods that parallel implement all operations at the edges of hyper networks, we integrate all candidate operators to a unified over-parameterized kernel. Fortunately, this kernel is originally from one of the candidate operations. Thus, the space complexity is greatly reduced by conducting parameter reusing. The unified over-parameterized kernel in our method is called ``hyper kernel'' to distinguish the ``super kernel'' in \cite{singlepathnas} and emphasize its abilities since it not only contains all candidate operations as the ``super kernel'' in \cite{singlepathnas}, but also can generate structural parameters. Naturally, our proposed one-shot NAS is named ``hyper kernel neural architecture search'' (HKNAS). Since the candidate operations of DARTS such as convolutions and poolings can not be integrated at the same time, instead of using the DARTS search space, we define a new search space similar to \cite{mobilenetv2} that contains hierarchical multiple blocks and layers to maintain structural diversity. In this search space, candidate operations only include standard convolutions to facilitate unified integrations. Based on this search space and HKNAS, we develop three different networks that are separately named 1-D HK-CLS, 3-D HK-CLS, and 3-D HK-SEG.  1-D HK-CLS and 3-D HK-CLS are used for pixel-level classification with spectral vectors or spatial patches. 3-D HK-SEG is used for image-level classification, which is also denoted as segmentation. What's more, when extracting spectral-spatial features, instead of using 3-D convolutions as previous methods \cite{3dcnn}, we adopt two sets of different convolutions in spectral and spatial aspects to substitute the 3-D convolution, and the flexibilities of network structures are further improved when combining the HKNAS. The main contributions of this paper can be summarized as follows:

\begin{itemize}

  \item[1)] We propose a novel architecture-searching solution called HKNAS. By analyzing the relationships between architectures and network weights, we propose to define structural parameters from hyper kernels, effectively reducing the searching difficulty by transforming the dual optimization problem into a one-tier optimization problem. Compared with existing NAS methods in the hyperspectral community, the proposed hyper kernels can also reuse the parameters of all candidate operations, drastically decreasing the model space complexity.
  \item[2)] We design a new search space for the convenience of HKNAS implementation, where hierarchical multiple blocks and layers are exploited to ensure architecture diversity. In addition, we separately use two kinds of convolutions to extract spectral and spatial information for substituting the conventional 3-D convolution, further enriching the network structure database.
  \item[3)] We develop three different networks named 1-D HK-CLS, 3-D HK-CLS, and 3-D HK-SEG on the foundation of the above search strategy and search space. These networks separately utilize spectral or spectral-spatial joint features for pixel or image-level HSI classification. Extensive quantitative and qualitative experiments demonstrate that the proposed methods outperform existing state-of-the-art approaches on six commonly used hyperspectral datasets.
\end{itemize}

The remainder of this paper is organized as follows. Section II introduces the related work. Section III describes the proposed HKNAS, the designed search space, and the final presented networks. Experiments and related comprehensive analyses are presented in Section IV. Finally, Section V concludes this paper.

\section{Related Work}

\subsection{Neural Architecture Search}

Early NAS attempts to control almost all components through RL or EA. The conception of NAS is given by NAS-RL \cite{NAS_RL}. It utilizes policy gradient optimization by regarding the architecture as a sequence. The MetaQNN \cite{metaqnn} that in the same period adopts the Q-learning algorithm, while Large-scale Evolution \cite{largescaleevo} and GeNet \cite{genetic_cnn} employ EA. However, because of enormous search spaces, these methods are difficult to find an optimal structure. Inspired by several CNN methods \cite{inception_v1,resnet} where repeated convolutional cells are utilized. NASNet \cite{nasnet} designs a network that stacks repeated convolutional motifs containing same architectures but with different weights. Thus, only the architecture inside cells needs to be searched, greatly simplifying the task difficulty. This design is followed by many subsequent works \cite{blockqnn,renas,amoebanet}. However, these methods still use discrete strategies such as RL or EA to search, requiring training and evaluating many architectures and causing unaffordable computational overheads. Then, the most famous gradient-based method DARTS \cite{darts} is turned up, which first develops a differentiable framework that facilitates the architecture searching through constantly relaxing structural parameters using the gradient algorithm. Then, many variants are developed to further improve it \cite{pdarts,pcdarts,sgas,snas}. Different from the above approaches, single-path NAS \cite{singlepathnas} fuses the parallel paths of candidate operations into an over-parameterized super kernel, where the selection of different operations is decided by the norm of kernel weights and individually defined learnable threshold values. Beyond architectures inside blocks, the latest methods \cite{networkadjustment,densenas,hrnas} have started searching macro structures such as block counts, block widths, and block connections in the search space of existing networks by considering resource costs. Compared with the above methods, although we still follow the idea of DARTS that uses structural parameters, in our method, the structural parameters are generated by hyper kernels instead of independently defining, strengthening the relationships between structural parameters and network weights.

\subsection{NAS for HSI classification}

In the HSI classification field, up to now, some methods have involved NAS mechanisms. Among these approaches, the first work of automatically designing the network structure is \cite{autocnn_hsi}, which directly employs DARTS to generate pixel-level classification networks named 1-D Auto-CNN and 3-D Auto-CNN, where the inputs are spectral vectors or spatial patches centered on target pixels, respectively. In 1-D Auto-CNN, spectral features are extracted by a series of candidate operations including 1-D convolutions and 1-D poolings. While in 3-D Auto-CNN, the authors apply 2-D instead of 3-D convolutions or poolings as candidate operations to obtain spectral-spatial features, although the network is called 3-D. Then, 3-D ANAS \cite{3danas} classifies the HSI based on an image-level classification network containing multiple cells, where the architecture inside each cell is generated by DARTS, while the outside connections between these cells are determined using viterbi algorithm \cite{viterbi}. In these cells, spectral-spatial features are produced by the candidate operations of 3-D asymmetric decomposition convolutions, where a 3-D convolution is substituted by a 1-D convolution and a 2-D convolution that have different kernel sizes. Besides ordinary convolutional operations, a recent pixel-level classification network SSTN \cite{sstn_nas} further considers attention modules, and the bilevel optimization algorithm for searching is separately carried out in two factorized subspaces. Compared with the above methods, we generate and evaluate both pixel-level and image-level classification networks on the foundation of the proposed HKNAS and the corresponding search space. In addition, although we also adopt two different convolutions to substitute the 3-D convolution, these two convolutions are practically two different operation families, which are separately derived from two different hyper kernels. These settings further improve the flexibility of network structures and reduce the number of trainable parameters.

\section{Proposed Method}
This section introduces the details of the proposed HKNAS and corresponding hierarchical multi-module search space. In HKNAS, the situations that use 1-D, 2-D, and 3-D convolutions as the candidate operations are discussed, respectively.

\subsection{Hyper Kernel Neural Architecture Search}

The convolution process can be seen as a pattern matching, where the areas in the input data whose patterns conform to filters will be highlighted, and asynchronous regions will be suppressed. Besides multiple convolutional kernels in conventional CNN are considered as abundant templates to separately recognize various patterns, the areas with different shapes and sizes inside kernels can also be regarded as different filters and used as sub kernels to represent different operations even if there is only one parent kernel. This conclusion has been demonstrated in \cite{singlepathnas}, where the parent kernel is called ``super kernel''.

Intuitively, in the hyper network of NAS procedures, the effectiveness of architectures highly relates to the expressiveness of features, which are generated by candidate operations, while these features possess high correlations with the operators of corresponding operations. Therefore, the searched architectures are implicitly related to the weights of candidate operations, while the architectures are determined by structural parameters. However, structural parameters are usually defined independently in previous methods, neglecting these relationships. To address this problem, we propose that structural parameters can be derived from network weights. 

Combining the above points, we can conclude that the structural parameters of candidate operations can be obtained from the sub kernels whose corresponding areas are included in parent kernels. Compared with the ``super kernel'' in \cite{singlepathnas}, the current parent kernels are not only able to contain multiple candidate operations but also can generate structural parameters. Thus, to distinguish from the previous ``super kernel'' and extra show their abilities, we call them ``hyper kernel''.

In this paper, to facilitate the integration of all operations into the hyper kernel and display the above thinking that generating structural parameters from sub kernels, we only consider multiple standard convolutions with different kernel sizes as candidate operations that can be derived from hyper kernels. All sub kernels are centered on the center pixel of the hyper kernel with a dilation rate of 1. These settings determine subsequent structural parameter generation procedures. In addition, we stipulate that the shape of sub kernels should be the same as their parent kernels, and the only difference is the kernel size.

\subsubsection{1-D and 2-D Convolution}

\begin{figure}[t]
  \centering
  \includegraphics[width=1\linewidth]{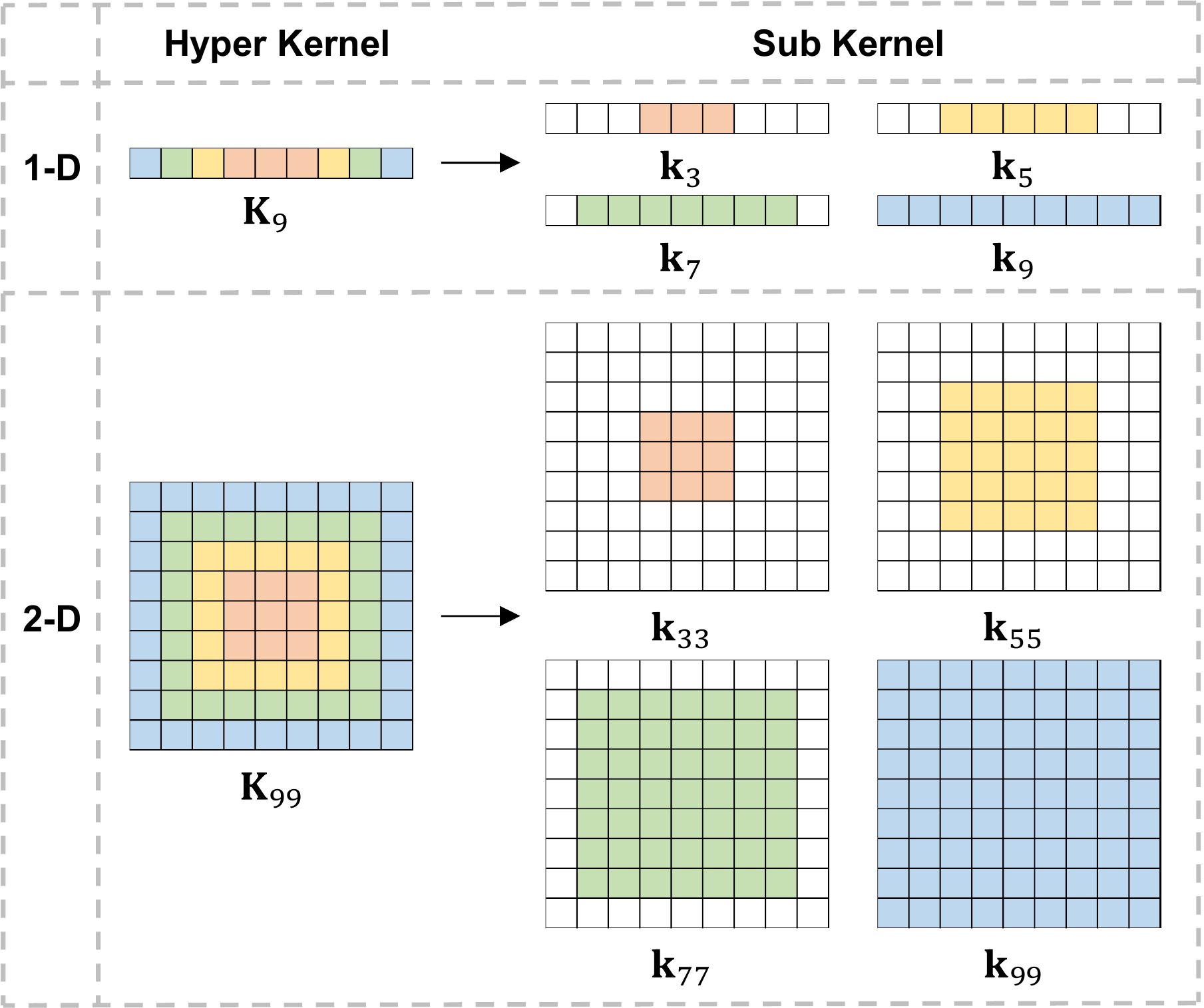}
  \caption{Hyper kernel and the correspondingly derived sub kernels in the example of 1-D and 2-D convolutions. Here, the sub kernels with different kernel sizes are marked with different colors and we show the situations where the kernel sizes of hyper kernels are 9 or 9 $\times$ 9, respectively.}
  \label{hyperkernel}
\end{figure}

Without loss of generality, Figure \ref{hyperkernel} shows the schematic diagrams of the hyper kernels that separately take 1-D and 2-D convolutional kernels as examples. The hyper kernels and sub kernels are separately symbolized as $\mathbf{K}$ and $\mathbf{k}$, where the number of subscripts is the number of dimensions in which the kernel shape can be changed, and the subscript values are the sizes of current hyper kernel or sub kernel. For the 1-D hyper kernel $\mathbf{K}_{S}$ in size of $1 \times S$ and 2-D hyper kernel $\mathbf{K}_{SS}$ in size of $S \times S$, where the $S$ is an odd number to ensure symmetry, the sub kernels what we obtain are shown as follows:
\begin{equation}
    \begin{split}
    \mathbf{K}_{S} \longrightarrow & \left\{\mathbf{k}_{(2s+1)} \Big| s=1,\cdots,\left\lfloor S/2\right\rfloor \right\} \\
    \mathbf{K}_{SS} \longrightarrow & \left\{\mathbf{k}_{(2s+1)(2s+1)}\Big| s=1,\cdots,\left\lfloor S/2\right\rfloor\right\} \\
  \end{split}
  \label{define_sub_kernel}
\end{equation}
Here, the smallest size of the convolutional kernel is 3 $\times$ 3 since we consider a 1 $\times$ 1 convolutional layer as the alias of a fully connected layer. Thus the $s$ begins from 1 instead of 0.

\begin{figure}[t]
  \centering
  \includegraphics[width=1\linewidth]{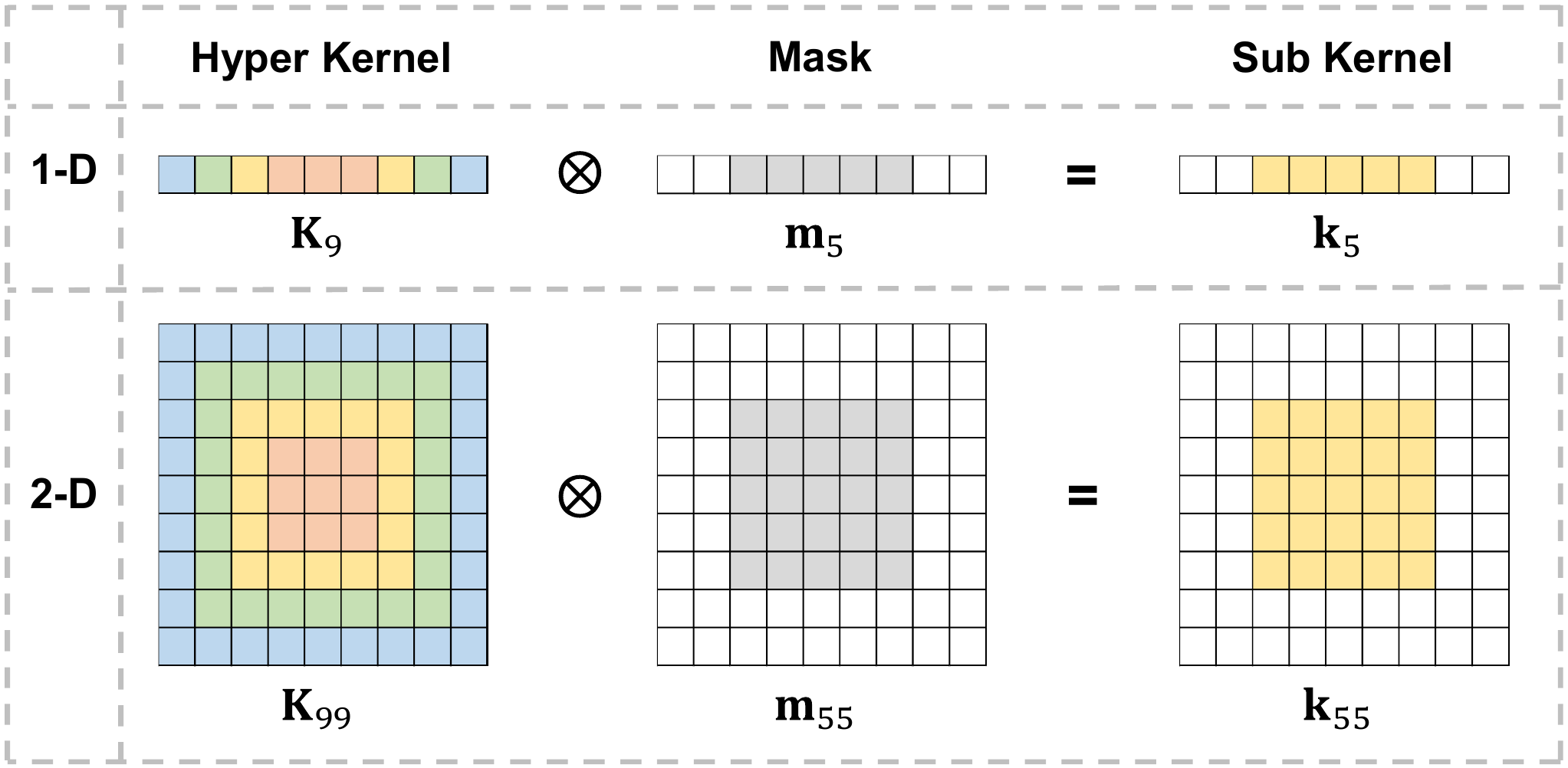}
  \caption{Procedures of generating sub kernels in size of 5 or 5 $\times$ 5 using binary masks in the example of 1-D and 2-D convolutions, respectively. Here, ``$\bigotimes$'' is element-wise multiplication.}
  \label{mask}
\end{figure}

To obtain sub kernels $\mathbf{k}$, we define a series of binary masks $\mathbf{m}$ where the implications of subscripts are the same as kernels. For each mask, the areas corresponding to its related kernels are set to 1, while the others are labeled as 0. Figure \ref{mask} displays the procedures of obtaining sub kernels in 1-D and 2-D situations, respectively. Thus, relationships between sub kernels and hyper kernels can be formulated as
\begin{equation}
  \begin{split}
  \mathbf{k}_{(2s+1)}^{i} &= \mathbf{K}_{S}^{i} \cdot \mathbf{m}_{(2s+1)}^{i} \\
  \mathbf{k}_{(2s+1)(2s+1)}^{ij} &= \mathbf{K}_{SS}^{ij} \cdot \mathbf{m}_{(2s+1)(2s+1)}^{ij} \\
\end{split}
\label{generate_sub_kernel}
\end{equation}
where 
\begin{equation}
  \begin{split}
    \mathbf{m}_{(2s+1)}^{i}&=
  \begin{cases}
   1&  \left\lfloor S/2\right\rfloor+1-s \leq i \leq \left\lfloor S/2\right\rfloor+1+s\\
   0& otherwise \\
\end{cases}\\
    \mathbf{m}_{(2s+1)(2s+1)}^{ij}&=
  \begin{cases}
   1&  \left\lfloor S/2\right\rfloor+1-s \leq i,j \leq \left\lfloor S/2\right\rfloor+1+s\\
   0& otherwise \\
\end{cases}
\label{define_mask}
  \end{split}
\end{equation}
Here, $\mathbf{m}_{(2s+1)}^{i}$ is the $i$th value of $\mathbf{m}_{(2s+1)}$, while $\mathbf{m}_{(2s+1)(2s+1)}^{ij}$ is the value of $\mathbf{m}_{(2s+1)(2s+1)}$ in $i$th row and $j$th column.

Since the necessary sub kernels all can be derived from hyper kernels. Compared with separately defining each sub kernel, directly utilizing a unified hyper kernel obviously requires fewer trainable parameters. After obtaining sub kernels, the next issue is how to produce structural parameters.

\begin{figure}[t]
  \centering
  \includegraphics[width=1\linewidth]{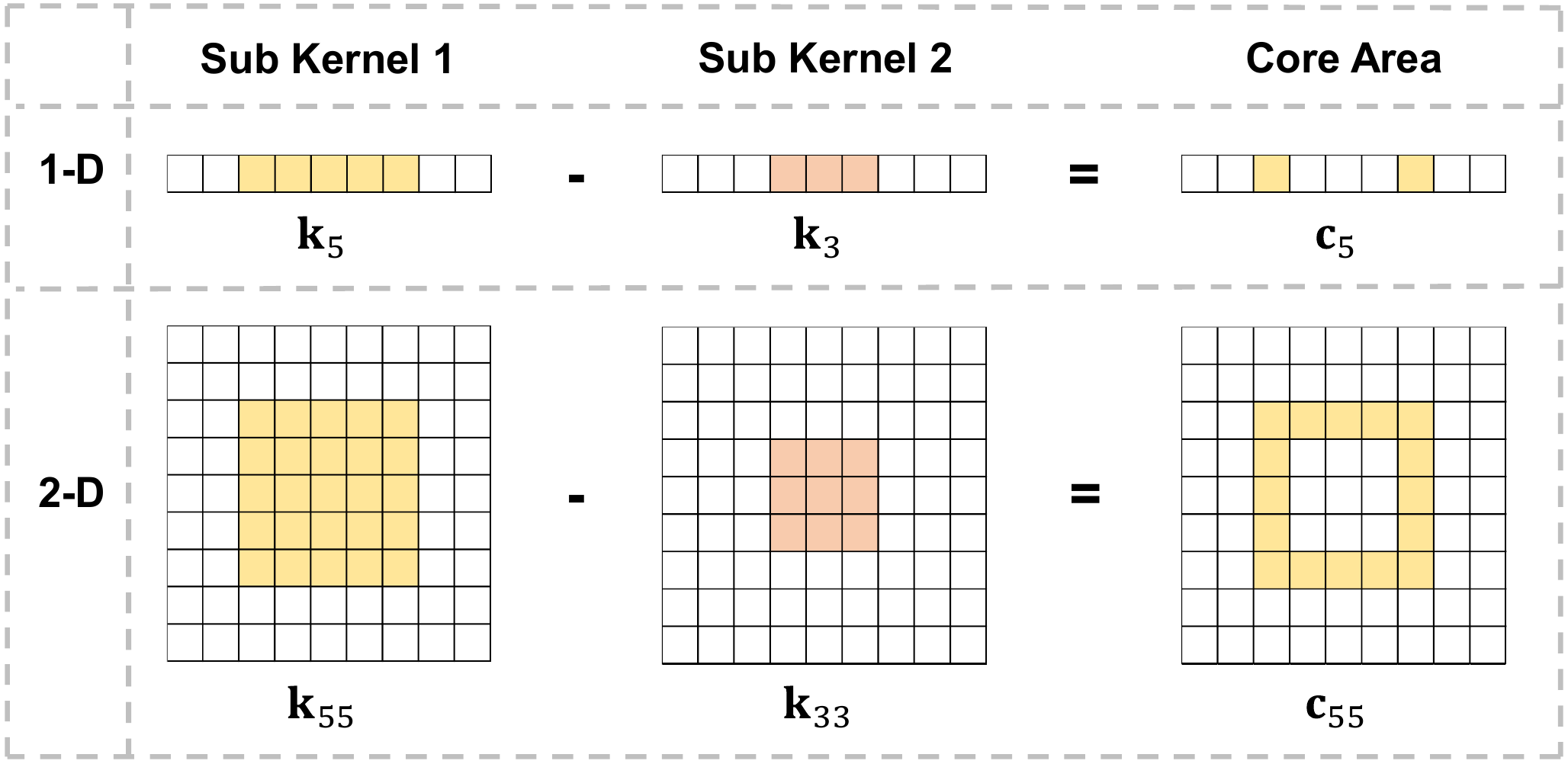}
  \caption{Procedures of generating core areas using two sub kernels separately in size of 5 and 3 or 5 $\times$ 5 and 3 $\times$ 3 in the example of 1-D and 2-D convolutions, respectively.}
  \label{corearea}
\end{figure}

From Figure \ref{hyperkernel}, it can be seen that the differences between sub kernel $\mathbf{k}_{33}$ and sub kernel $\mathbf{k}_{55}$ are the pixels in the outermost circle. Thus, to measure the impact of $\mathbf{k}_{55}$, we only need to pay attention to this annular area. To this end, we define a concept named core area for each sub kernel to compute corresponding structural parameters. Figure \ref{corearea} depicts the illustrations of core areas to clearly explain this concept. Especially, the core area of the sub kernel with the smallest kernel size is itself. Thus, the core areas of all derived sub kernels can be represented as
\begin{equation}
  \begin{split}
    & \mathbf{c}_{(2s+1)}=
  \begin{cases}
  \mathbf{k}_{(2s+1)}&  s=1\\
  \mathbf{k}_{(2s+1)}-\mathbf{k}_{(2s-1)}& 1<s\leq \left\lfloor S/2\right\rfloor\\
\end{cases}\\
    & \mathbf{c}_{(2s+1)(2s+1)}= \\
  &\begin{cases}
  \mathbf{k}_{(2s+1)(2s+1)}&  s=1\\
  \mathbf{k}_{(2s+1)(2s+1)}-\mathbf{k}_{(2s-1)(2s-1)}& 1<s\leq \left\lfloor S/2\right\rfloor\\
\end{cases}\\
  \end{split}
  \label{core_area}
\end{equation}
Correspondingly, the masks of core areas are
\begin{equation}
  \begin{split}
    & \mathbf{m}_{(2s+1)}^{(c)}=
  \begin{cases}
  \mathbf{m}_{(2s+1)}&  s=1\\
  \mathbf{m}_{(2s+1)}-\mathbf{m}_{(2s-1)}& 1<s\leq \left\lfloor S/2\right\rfloor\\
\end{cases}\\
    & \mathbf{m}_{(2s+1)(2s+1)}^{(c)}= \\
  &\begin{cases}
  \mathbf{m}_{(2s+1)(2s+1)}&  s=1\\
  \mathbf{m}_{(2s+1)(2s+1)}-\mathbf{m}_{(2s-1)(2s-1)}& 1<s\leq \left\lfloor S/2\right\rfloor\\
\end{cases}\\
  \end{split}
  \label{core_area_mask}
\end{equation}

Then, the structural parameter of each candidate operation can be computed from the core area of the corresponding sub kernel. In this paper, we calculate the structural parameter by averaging the weight values of the core area since the magnitude of the weight values in the core area indicates the relative importance of candidate operations. Thus, the obtained structural parameters are
\begin{equation}
  \begin{split}
  \alpha_{s}^{(1)} &= \frac{\sum \mathbf{c}_{(2s+1)}}{\sum \mathbf{m}_{(2s+1)}^{(c)}} \\
  \alpha_{s}^{(2)} &= \frac{\sum \mathbf{c}_{(2s+1)(2s+1)}}{\sum \mathbf{m}_{(2s+1)(2s+1)}^{(c)}} \\
\end{split}
\label{structural_parameter}
\end{equation}

In fact, from the settings of hyper kernels and sub kernels it can be seen that the number of candidate operations is highly related to the size of the hyper kernel, and surely is $\left\lfloor S/2\right\rfloor$. Assume all candidate operations $o^{(1)}$ and $o^{(2)}$ that are separately derived from hyper kernels $\mathbf{K}_{S}$ and $\mathbf{K}_{SS}$ belong to set $\mathcal{O}^{(1)}$ and $\mathcal{O}^{(2)}$, respectively. The capacity of $\mathcal{O}$ is $|\mathcal{O}^{(1)}|=|\mathcal{O}^{(2)}|=\left\lfloor S/2\right\rfloor$. Following \cite{darts}, the edge prepared to be determined in the searching stage of the hyper network in the proposed HKNAS is
\begin{equation}
  f^{(d)}\left(\mathbf{x}^{(d)}\right) = \sum_{s=1}^{\left|\mathcal{O}^{(d)}\right|} \frac{\exp\left(\alpha_{s}^{(d)}\right)}{\sum_{s'=1}^{\left|\mathcal{O}^{(d)}\right|} \exp\left(\alpha_{s'}^{(d)}\right)} o_s^{(d)} \left(\mathbf{x}^{(d)}\right) \\
  \label{feed_forward}
\end{equation}
Here, $d$ is the dimension number that can be changed in the hyper kernel. For example, $\mathbf{x}^{(2)}$ is a 2-D input feature in size of $C \times H \times W$, where $C$, $H$ and $W$ are separately the channel number, height and width.

\begin{algorithm}[t]
  \caption{HKNAS-Hyper Kernel Architecture Search}
  \KwData{training set $\mathcal{D}_{trn}$, learning rate $\xi$, binary masks $\mathbf{m}$}
  \KwResult{The learned $\alpha^{(d)}$}
  Randomly initialize hyper kernels $\mathbf{K}$ from $\mathcal{N}(0,1)$\\ \label{hknas_algorithm}
  \While{not converged}{
  Generate sub kernels $\mathbf{k}$ by $\mathbf{K}$ and $\mathbf{m}$\;
  Obtain operation set $\mathcal{O}^{(d)}$ parameterized by $\mathbf{k}$\;
  Compute structural parameters $\alpha^{(d)}$ by $\mathbf{k}$ and $\mathbf{m}$\;
  Construct mixed operation $f^{(d)}$\; 
  Training on $\mathcal{D}_{trn}$ and obtain loss $\mathcal{L}_{trn}(\mathbf{K})$\;
  $\mathbf{K} \leftarrow \mathbf{K} - \xi\nabla_{\mathbf{K}}\mathcal{L}_{trn}(\mathbf{K})$\;}
  \end{algorithm}

  Algorithm \ref{hknas_algorithm} presents the workflow of HKNAS using one-tier optimization. Here, the subscript $S$ and $SS$ are omitted for convenience. After searching, we select the operation $o_{s^*}^{(d)}$ with the largest structural parameter, where
\begin{equation}
  s^* = \arg\max\limits_{s} \alpha_s^{(d)}
  \label{index_12d}
\end{equation}

\subsubsection{3-D Convolution}

Obviously, the above procedures for 1-D or 2-D hyper kernels can be easily extended to 3-D situations to obtain $\mathbf{K}_{SSS}$ and $\mathcal{O}^{(3)}$. However, 3-D convolution usually needs a large number of trainable parameters and costs too many computational resources. To alleviate this problem, \cite{3danas} splits the 3-D convolution into a 1-D convolution and a 2-D convolution for spectral and spatial information perception, respectively. Inspired by their work, we further combine this thinking with our HKNAS, generating more abundant network structures.

Before introducing our method, we first review the formula of 3-D convolution, for an input 3-D feature $\mathbf{x}$ in size of $C \times D \times H \times W$, where $D$ is the depth (the meanings of $C, H, W$ are the same as 2-D features and we omit the superscript $(3)$ for convenience), we have
\begin{equation}
    \mathbf{z} = \sum\limits_{h=1}^{p}  \sum\limits_{w=1}^{p}  \sum\limits_{t=1}^{q} 
    \mathbf{W}_{hwt} \cdot \mathbf{x}_{(i+h-\left\lceil p/2\right\rceil)(j+w-\left\lceil p/2\right\rceil)(k+t-\left\lceil q/2\right\rceil)}
    \label{3d}
\end{equation}
where the 3-D convolutional kernel $\mathbf{W}$ is in size of $p \times p \times q$ ($p,q$ are all odd numbers), ``*'' is 3-D convolution operation. Formula \ref{3d} can be rewritten as 
\begin{equation}
  \begin{split}
  \mathbf{y}_{ij} = &\sum\limits_{t=1}^{q} \mathbf{K}_{q}^t \cdot \mathbf{x}_{(i)(j)(k+t-\left\lceil q/2\right\rceil)}\\
  \mathbf{z} = & \sum\limits_{h=1}^{p}  \sum\limits_{w=1}^{p} \mathbf{K}_{pp}^{(dw),hw} \mathbf{y}_{(i+h-\left\lceil p/2\right\rceil)(j+w-\left\lceil p/2\right\rceil)}\\
  \end{split}
  \label{1d_2ddw}
\end{equation}
It can be seen that a 3-D convolution can be decomposed by a 1-D convolution and a 2-D depth-wise convolution. Here, to distinguish from the standard convolution, the hyper kernel of depth-wise convolution is marked by a superscript $(dw)$. The corresponding operations of the sub kernels that are from depth-wise convolutional hyper kernels are still depth-wise convolutions. Since all hyper kernels can be substituted by their derived sub kernels, two kinds of candidate operation sets are naturally used. It should be noted that, in our implementation, $p=q$. Thus, there are $\left\lfloor p/2\right\rfloor ^2 $ architectures in total, greatly improving the network flexibility compared with only adopting 3-D hyper kernels, which only contains $\left\lfloor p/2\right\rfloor$ options. In the current hyper network, the corresponding procedures are
\begin{equation}
  f(\mathbf{x})=f^{(2,dw)}\left(f^{(1)}(\mathbf{x})\right)
\end{equation}
where 
\begin{equation}
  \begin{split}
  f^{(1)}\left(\mathbf{x}\right) = & \sum_{s=1}^{\left\lfloor p/2\right\rfloor} \frac{\exp\left(\alpha_{s}^{(1)}\right)}{\sum_{s'=1}^{\left\lfloor p/2\right\rfloor} \exp\left(\alpha_{s'}^{(1)}\right)} o_s^{(1)} \left(\mathbf{x}\right) \\
  f^{(2,dw)}\left(\mathbf{x}\right) = & \sum_{r=1}^{\left\lfloor p/2\right\rfloor} \frac{\exp\left(\alpha_{r}^{(2,dw)}\right)}{\sum_{r'=1}^{\left\lfloor p/2\right\rfloor} \exp\left(\alpha_{r'}^{(2,dw)}\right)} o_r^{(2,dw)} \left(\mathbf{x}\right) \\
  \end{split}
\end{equation}
Here, $\alpha^{(2,dw)}$ and $o^{(2,dw)}$ are separate structural parameters and corresponding candidate operations of the sub kernels that are derived from the hyper kernel using for 2-D depth-wise convolution, while $f^{(2,dw)}$ is the edge that all candidate operations are 2-D depth-wise convolutions. Also, for the searched operations $o_{s^*}^{(1)}$ and $o_{r^*}^{(2,dw)}$, we have

\begin{equation}
  \begin{split}
  s^* = &\arg\max\limits_{s} \alpha_s^{(1)}\\
  r^* = &\arg\max\limits_{r} \alpha_r^{(2,dw)}\\
  \end{split}
  \label{index_3d}
\end{equation}

In fact, the order of 1-D convolution and 2-D depth-wise convolution can be exchanged. The concrete configurations will be determined in later experiments.

\subsection{Framework of Search Space}

\begin{figure}[t]
  \centering
  \includegraphics[width=1\linewidth]{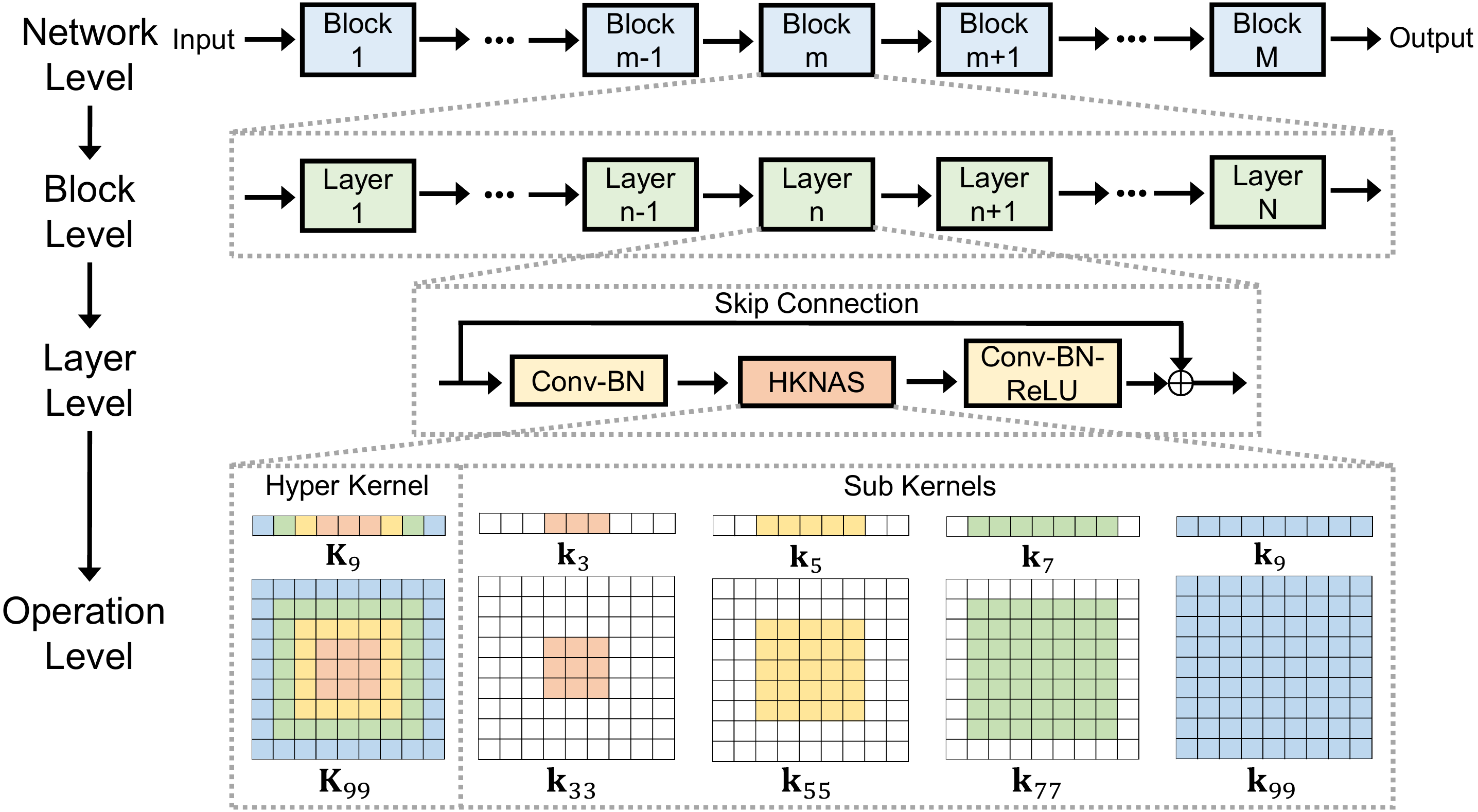}
  \caption{The overall network framework includes four levels. The whole network is composed of multiple blocks, where each block contains many layers. In each layer, three parts are included. Among these modules, the second part is the architecture needed to be searched by HKNAS, in which the hyper kernels that can derive sub kernels for candidate operations are included.}
  \label{framework}
\end{figure}

Since we only adopt the standard convolutional layer as the candidate operation, in this paper, instead of employing the DARTS search space used in \cite{autocnn_hsi} that contains pooling operations that are difficult to be integrated into hyper kernels, we develop a simple but effective new search space for the proposed HKNAS. In this search space, hierarchical multiple blocks and layers are included to improve the network diversity.

Figure \ref{framework} presents the whole framework of this search space. Following two popular frameworks \cite{resnet, mobilenetv2}, the whole network is designed to be formed by stacking multiple blocks. In each block, multiple layers are stacked to conduct feature extraction, where the feature resolutions are always maintained. Each layer inside blocks is a residual module, which is similar to the bottleneck in \cite{resnet} and can be regarded as three parts. The first and last parts are implemented with 1 $\times$ 1 convolutions to separately squeeze and recover the number of channels to one quarter and four times, so as to reduce computations, performing information integration while not changing channel numbers. Following \cite{mobilenetv2}, there is no rectified linear unit (ReLU) activation in the first part since ReLU removes negative values and may decrease the representation capability of the features, affecting the perception effect of subsequent operations. The input features of the current layer are added with the outputs through a skip connection for the convenience of gradient propagation, so as to accelerate convergence. The most important part is the second part, which is determined by the proposed HKNAS, where the hyper kernel that contains sub kernels involving all candidate operations and can generate structural parameters is adopted. For convenience, the category of the candidate operation set in one layer is the same as other layers even if in different blocks, meaning a network only has one kind of candidate operation set. Nevertheless, different from the modular-based search space in \cite{darts} that directly stacks cells of the same architecture, we separately determine the architecture for each layer, meaning the whole framework is more diverse and flexible since it possesses a large search space, and the structure of the network can be exquisitely controlled.

To improve the generality of this framework. We don't fix the network depth, while the optimal number of blocks $M$ and the number of layers $N$ inside one block will be determined in later experiments to obtain better network structures.

\section{Experiments}

This section presents a series of comprehensive qualitative and quantitative analyses of the proposed methods on six public scenarios in the HSI classification community.

\subsection{Dataset}

Besides four classical HSI classification benchmarks, we also use two complex scenes of a recently proposed challenging WHU-Hi dataset \cite{whuhi} to evaluate the proposed methods.
      
  \begin{itemize}
    \item[1)] Indian Pines: This scene was gathered at North-western Indiana by Airborne Visible/Infrared Imaging Spectrometer (AVIRIS) sensor in 1992, consisting 200 bands with size of 145 $\times$ 145 pixels that in 20m spatial resolution after water absorption bands were removed and in the wavelength range of 0.4-2.5$\mu$m. 16 vegetation classes are involved in this scene.
    \item[2)] Pavia University: This scene was obtained over Pavia University at Northern Italy by Reflective Optics System Imaging Spectrometer (ROSIS) in 2001, consisting 103 bands with size of 610 $\times$ 340 pixels that in 1.3m spatial resolution and in the wavelength range of 0.43-0.86$\mu$m. 9 categories are included in this data.
    \item[3)] Kennedy Space Center: This scene was collected at Kennedy Space Center, Florida by AVIRIS in 1996. After removing water absorption and low SNR bands, 176 bands in range of 0.4-2.5$\mu$m are remained. 512 $\times$ 614 pixels are included in this dataset and in spatial resolution of 18m. 13 classes representing the various land cover types that occur in this environment were defined for classification. 
    \item[4)] Salinas Valley: This scene was collected by AVIRIS sensor over Salinas Valley, California, consisting 204 bands with size of 217 $\times$ 512 samples that in 3.7m spatial resolution. As with Indian Pines scene, 20 water absorption bands were discarded. It contains 16 classes, including vegetables, bare soils, and vineyard fields.
    \item[5)] WHU-Hi: It is a recently released challenging dataset acquired by Wuhan University with unmanned aerial vehicle-borne hyperspectral systems in Hubei province, China. It has three scenes including LongKou, HanChuan, and HongHu. In this paper, we utilize two complex agriculture scenarios: HanChuan and HongHu. They separately contain 274 and 270 bands in the wavelength range of 0.4-1.0$\mu$m, and have 16 and 22 classes, respectively. 1217 $\times$ 303 and 940 $\times$ 475 pixels are separately included in these scenes.  
  \end{itemize}

To better indicate the effectiveness of the proposed methods, we only use a few samples for searching and training. Thus, besides the Grass-pasture-mowed and Oats categories in the Indian Pines scene separately adopt 14 and 10 samples for training and validation (training set and validation set each account for half). The remained classes in this scene and the categories of other datasets all possess 20 knowable samples. 

\subsection{Implementation Details}

\begin{table}[t]
  \caption{Different settings of the developed three kinds of networks}
  \newcommand{\tabincell}[2]{\begin{tabular}{@{}#1@{}}#2\end{tabular}}
  \centering
  \label{scheme_222_842}
  \resizebox{\linewidth}{!}{
  \begin{tabular}{|l|l|l|l|}
  \hline
  Network & Input & Candidate operation & Output  \\
  \hline
  1-D HK-CLS &\tabincell{l}{Spectral vectors of \\ target pixel} &  1-D Convolutions & Category \\
  \hline
  3-D HK-CLS &\tabincell{l}{Patches centered on \\ target pixel} &  3-D Convolutions & Category \\
  \hline
  3-D HK-SEG &\tabincell{l}{Image Patches or \\ Whole image} & 3-D Convolutions & Classification map  \\
  \hline
\end{tabular}
  }
  \label{three_networks}
\end{table}

\subsubsection{Model Details}

Leveraging the proposed search strategy HKNAS and the search space of hierarchical multiple modules, we search three kinds of networks that separately for pixel-level classification using spectral vectors, pixel-level classification using spatial patches centered on target pixel, and image-level classification. These three networks are called 1-D HK-CLS, 3-D HK-CLS, and 3-D HK-SEG. They are mainly different in the configurations of input, output, and the candidate operations of the HKNAS part (corresponding to different hyper kernels). The setting comparisons of them have been shown in Table \ref{three_networks}. We use ``SEG'' as a name since the image-level HSI classification network is similar to the segmentation network adopted in the computer vision field.

\begin{itemize}
  \item [1)] 1-D HK-CLS: This network is named ``1-D'' since it is a 1-D CNN that uses 1-D data to extract spectral information. The input spectral vectors of the normalized original image first pass through a fully connected layer to transform the feature length to 96, so as to reduce computations. Then multiple blocks are used to extract deep features, which are then merged using a 1-D average pooling and flattened through another fully connected layer for classification. The feature sizes are downsampled three times with $M/4$ blocks as intervals, where $M$ is the number of blocks, and the number of feature channels is doubled using a 1 $\times$ 1 convolution after each downsampling. The initial channel number of the first block is 64. In HKNAS parts, there are only 1-D hyper kernels that include 1-D sub kernels, whose corresponding candidate operations are all 1-D convolutions, after which a BN layer and a ReLU function are appended.
  \item [2)] 3-D HK-CLS: This network also implements pixel-level classification. Different from 1-D HK-CLS, it receives the spatial patches centered on the target pixel and uses ``3-D convolution'' to extract spectral-spatial features. The channel number of inputs is first transformed to 64 with a 1 $\times$ 1 convolution. Since the size of input spatial patches is set to 27 $\times$ 27, 3-D HK-CLS only adopts three blocks and the feature resolutions are downsampled after each block, while the number of channels is doubled using a 1 $\times$ 1 convolution in the mean time. A 2-D average pooling is adopted on the output of the last block, which is then flattened to pass through a fully connected layer for classification. In HKNAS parts, as in previous descriptions, the ``3-D convolution'' may be replaced by a 1-D convolution and a 2-D depth-wise convolution, which are searched by separately utilizing 1-D and 2-D hyper kernels. The BN layer and the ReLU layer are only added to the features that are processed by standard or approximate 3-D convolutions.
  \item [3)] 3-D HK-SEG: Different from the above two networks, this network conducts image-level classification where the whole image can be used as an input. Same as 3-D HK-CLS, it first maps the channel number to 64 through a 1 $\times$ 1 convolutional layer. To maintain high resolutions, features are downsampled only once, and the pooling layer is arranged on the location after the first block, while the number of channels for each block is doubled compared with the prior block. This is similar to the dilated FCN whose output stride equals 8 in the computer vision field. The outputs from the last block are directly upsampled using a bilinear interpretation function to generate the final classification map. It should be noted that the BN layers in 1-D HK-CLS and 3-D HK-CLS are substituted by the group normalization layer since the batch size in 3-D HK-SEG is 1. The configurations of HKNAS parts are totally the same as 3-D HK-CLS.
\end{itemize}

\subsubsection{Experimental Settings}

To perform the HSI classification, the whole procedure of our methods is conducted with three stages: searching stage, training stage, and evaluation stage, while the used dataset is also randomly split into three parts: training set, validation set, and testing set. Specifically, we first implement the searching stage, where the obtained network structures are then used to be trained from scratch. In the end, we evaluate the trained networks. Since we set the kernel sizes of hyper kernels to be 9 or 9 $\times$ 9 in 1-D or 2-D convolutions. Thus, as Figure \ref{hyperkernel} shows, there are four kinds of sub kernels separately in 1-D or 2-D hyper kernels, and the capacity of each candidate operation set is 4. In 1-D hyper kernels, the candidate operation set $\mathcal{O}^{(1)}$ contains: \textit{1 $\times$ 3 convolution}, \textit{1 $\times$ 5 convolution}, \textit{1 $\times$ 7 convolution} and \textit{1 $\times$ 9 convolution}. Similarly, \textit{3 $\times$ 3 $\times$ 3  convolution}, \textit{5 $\times$ 5 $\times$ 5 convolution}, \textit{7 $\times$ 7 $\times$ 7 convolution} and \textit{9 $\times$ 9 $\times$ 9 convolution} are considered as the candidate operations in $\mathcal{O}^{(3)}$ if 3-D hyper kernels are adopted. While \textit{3 $\times$ 3 depth wise convolution}, \textit{5 $\times$ 5 depth wise convolution}, \textit{7 $\times$ 7 depth wise convolution} and \textit{9 $\times$ 9 depth wise convolution} are included in candidate operation set $\mathcal{O}^{(2)}$ to simulate spatial information perceptions, which can be used to combine with 1-D convolutions, so as to substitute the standard 3-D convolutions in 3-D HK-CLS and 3-D HK-SEG.

In the searching stage, different from previous methods that separately optimize network weights and structural parameters using the training set and the validation set. We directly update both of them with the training set, while the validation set is used to monitor the network status. The learning rate and weight decay are both set to 0.01 and the learning rate is constantly scheduled with the cosine annealing function $current\_lr = min\_lr + \cfrac{1}{2}(initial\_lr-min\_lr)(1+\cos(\frac{current\_epoch}{max\_epoch}))$, where $min\_lr$ is 0. Since the 3-D HK-SEG possesses a large number of trainable parameters and receives more data, compared with 1-D HK-CLS and 3-D HK-CLS that adopt stochastic gradient descent algorithm, we extra introduce the momentum, which is set to 0.9, to better jump out of local optimization and accelerate convergence. The batch size in 1-D HK LS and 3-D HK-CLS is 96, while the same coefficient in 3-D HK-SEG is 1 since the whole image is fed into the model. The epoch of 1-D HK-CLS is 600, while 3-D HK-CLS and 3-D HK-SEG are configured as 100. We adopt the cross entropy loss for classification.

In the training stage, besides epochs, which are 1000 in 1-D HK-CLS and 300 in 3-D HK-CLS and 3-D HK-SEG, the other hyperparameters such as learning rate, batch size, loss function, and so on are totally the same as the searching stage. Also, the network weights of the searched architectures are updated only with the training set. In the end, the networks after training are evaluated with the testing set.

All experiments are repeatedly conducted 10 times with Pytorch framework on the Intel Xeon Gold 5118 2.30GHz processor and a single NVIDIA Tesla V100 GPU. Three commonly used quantitative evaluation criterions in the HSI classification community including overall accuracy (OA), average accuracy (AA), and kappa coefficient (Kappa) are applied, which are recorded as the mean value $\mu$ and standard deviation $\sigma$. The OA is the most popular evaluation index, which can be calculated through dividing the number of correct pixels by the number of all classified pixels. However, the OA is seriously affected by unbalance categories. To tackle this problem, AA and Kappa are separately computed based on the confusion matrix. The AA is obtained by averaging the recall values of all categories, while we use Kappa to measure the classification consistency to penalize the model that possesses category preferences.

The subsequent experiments are arranged as follows. Taking the classical HSI datasets as examples, we first show the process of hyperparameter determinations and component ablation studies to illustrate how to promote the above three kinds of networks to find optimal architectures. Then, the comparisons between the searched networks and existing state-of-the-art NAS-related methods for HSI classification are presented. At last, a series of analyses including model complexity discussions, loss curves, and the visualization of the searched architectures are performed to better understand the proposed methods.

\subsection{Hyper Parameter Analysis}

\begin{figure}[t]
  \centering
  \subfigure[]{\includegraphics[width=0.49\linewidth]{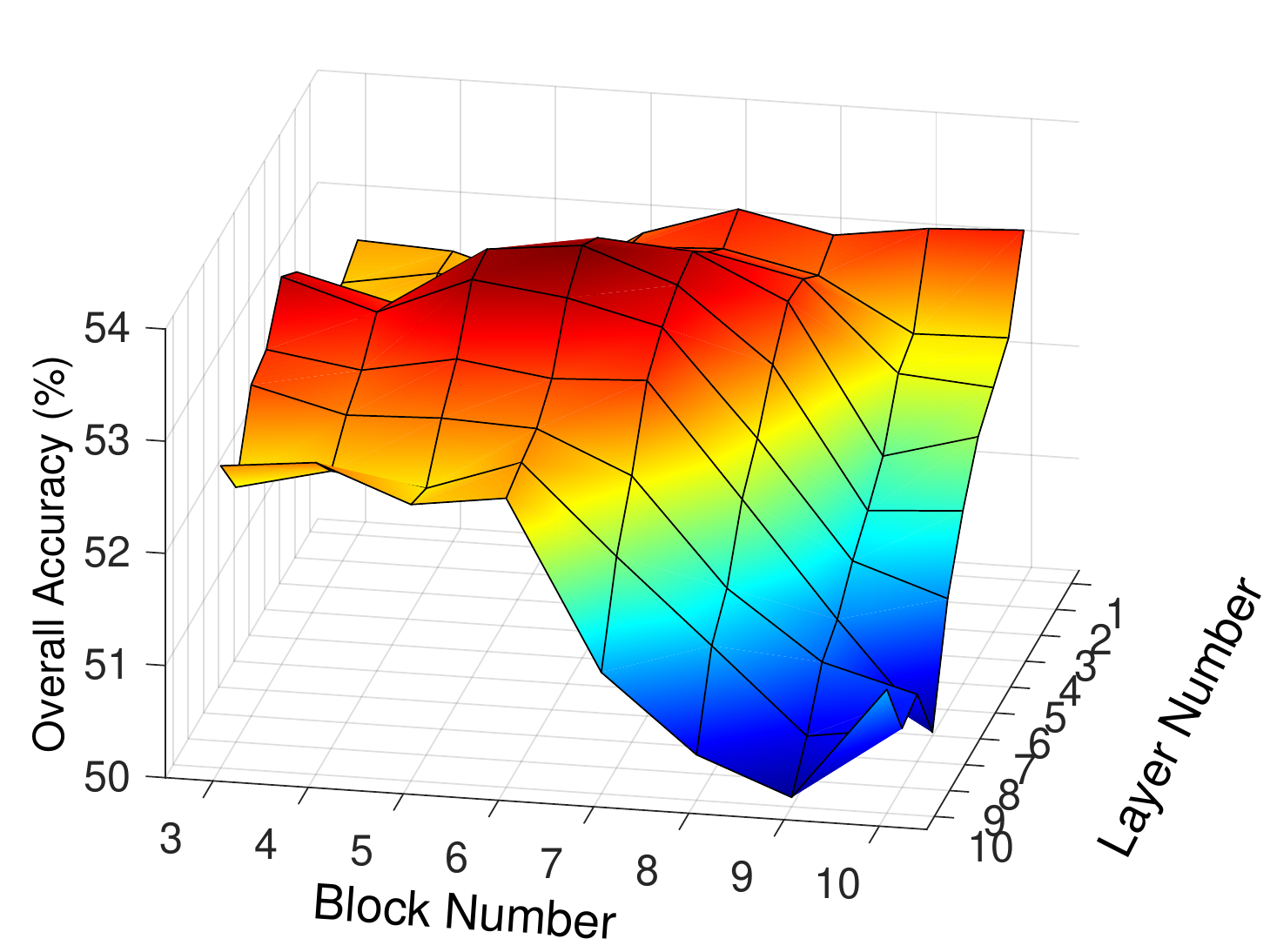}}
  \subfigure[]{\includegraphics[width=0.49\linewidth]{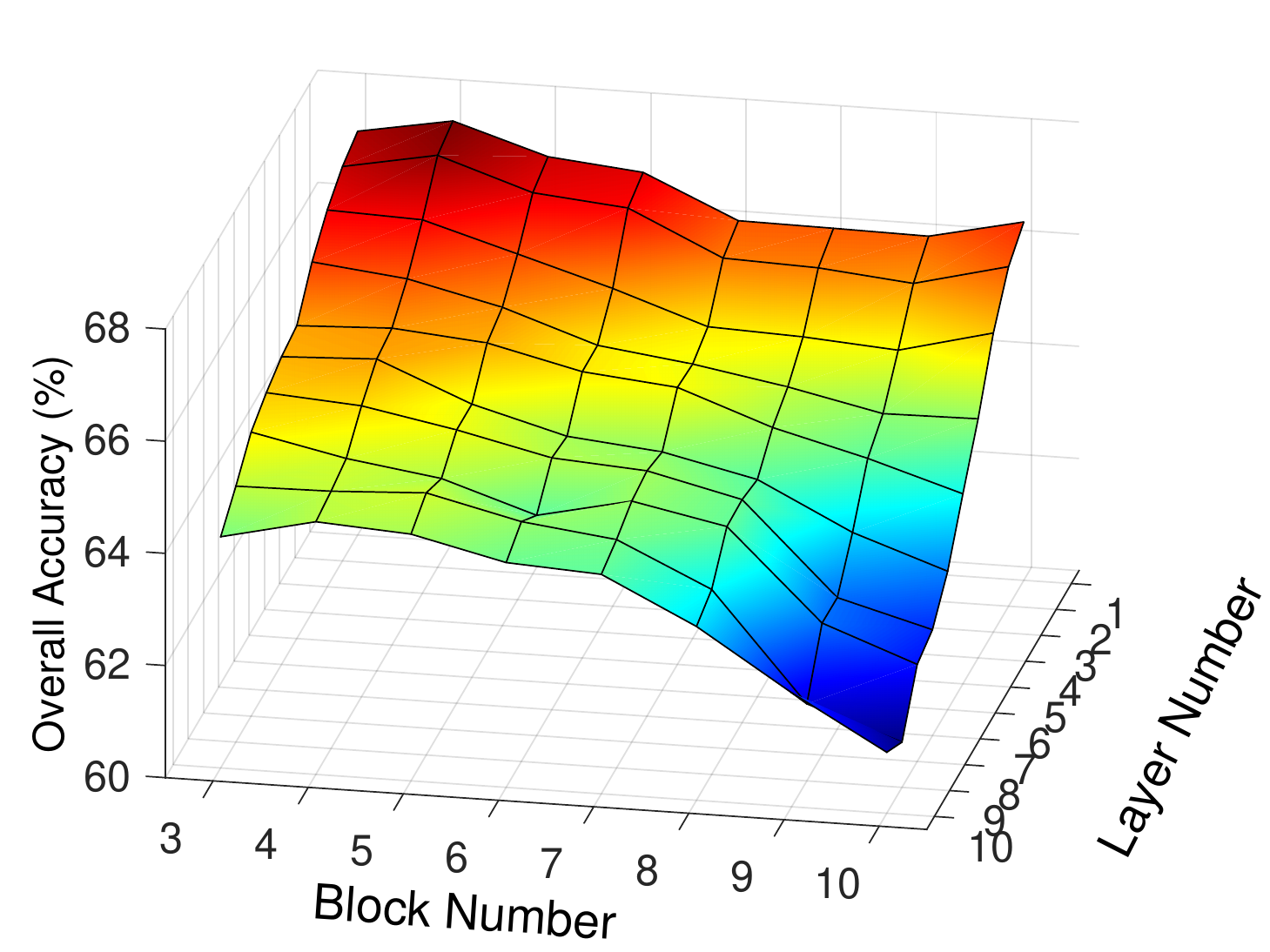}}
  \subfigure[]{\includegraphics[width=0.49\linewidth]{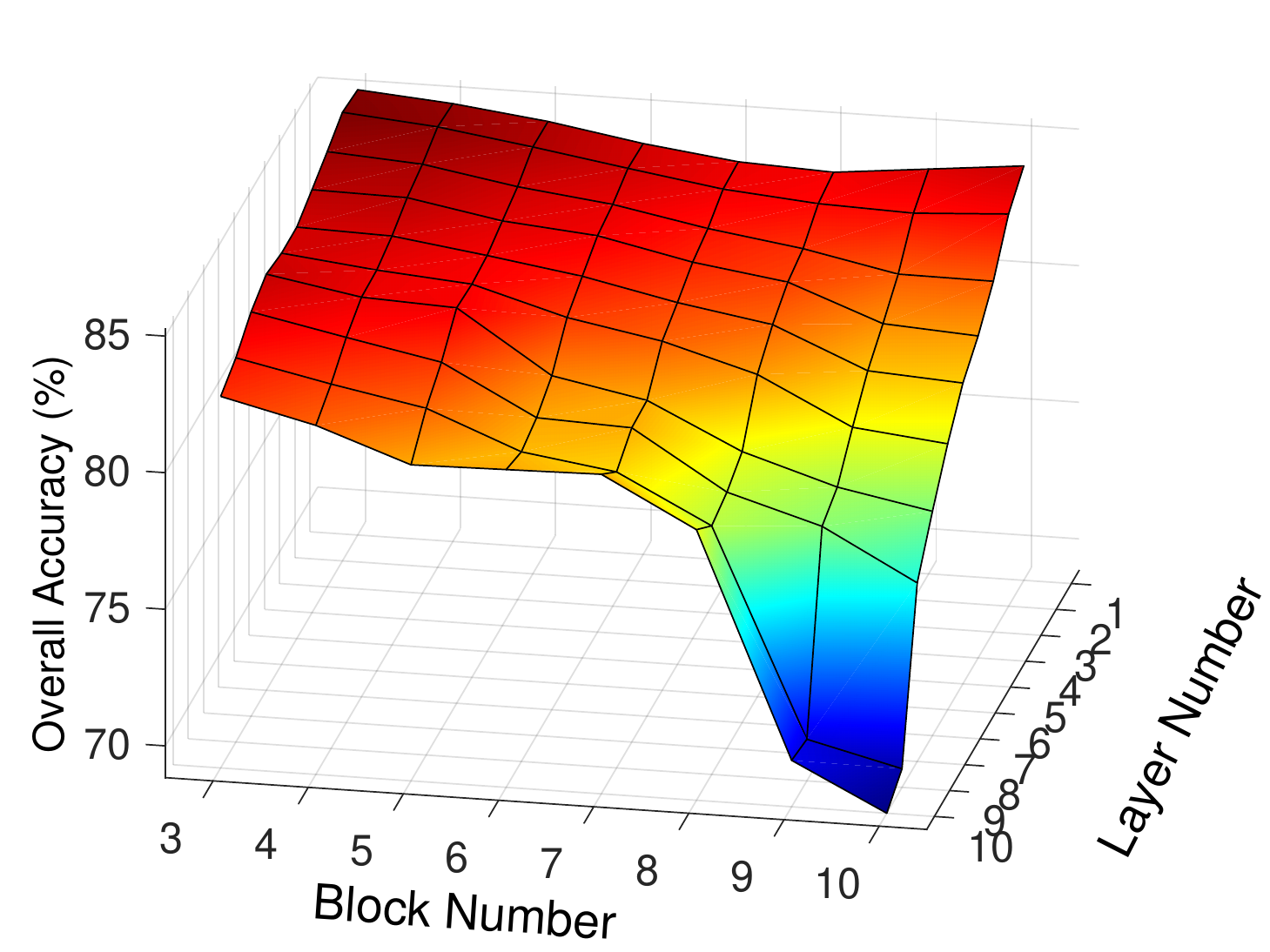}}
  \subfigure[]{\includegraphics[width=0.49\linewidth]{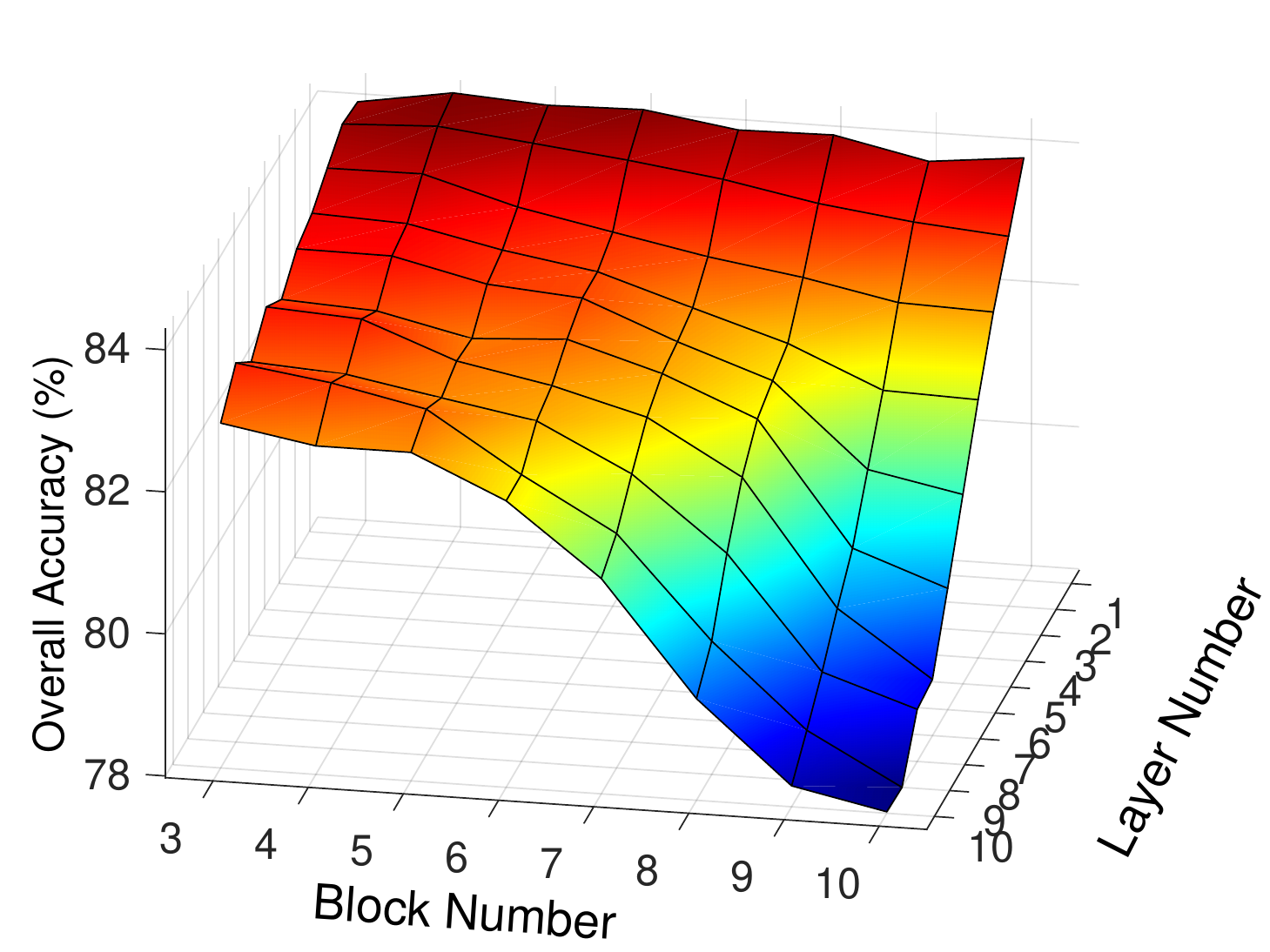}}
  \caption{Relationships between the block number, layer number, and the overall accuracy for 1-D HK-CLS on different datasets. (a) Indian Pines. (b) Pavia University. (c) Kennedy Space Center. (d) Salinas Valley.}
  \label{block_layer_acc}
\end{figure}

\begin{figure}[t]
  \centering
  \subfigure[]{\includegraphics[width=0.49\linewidth]{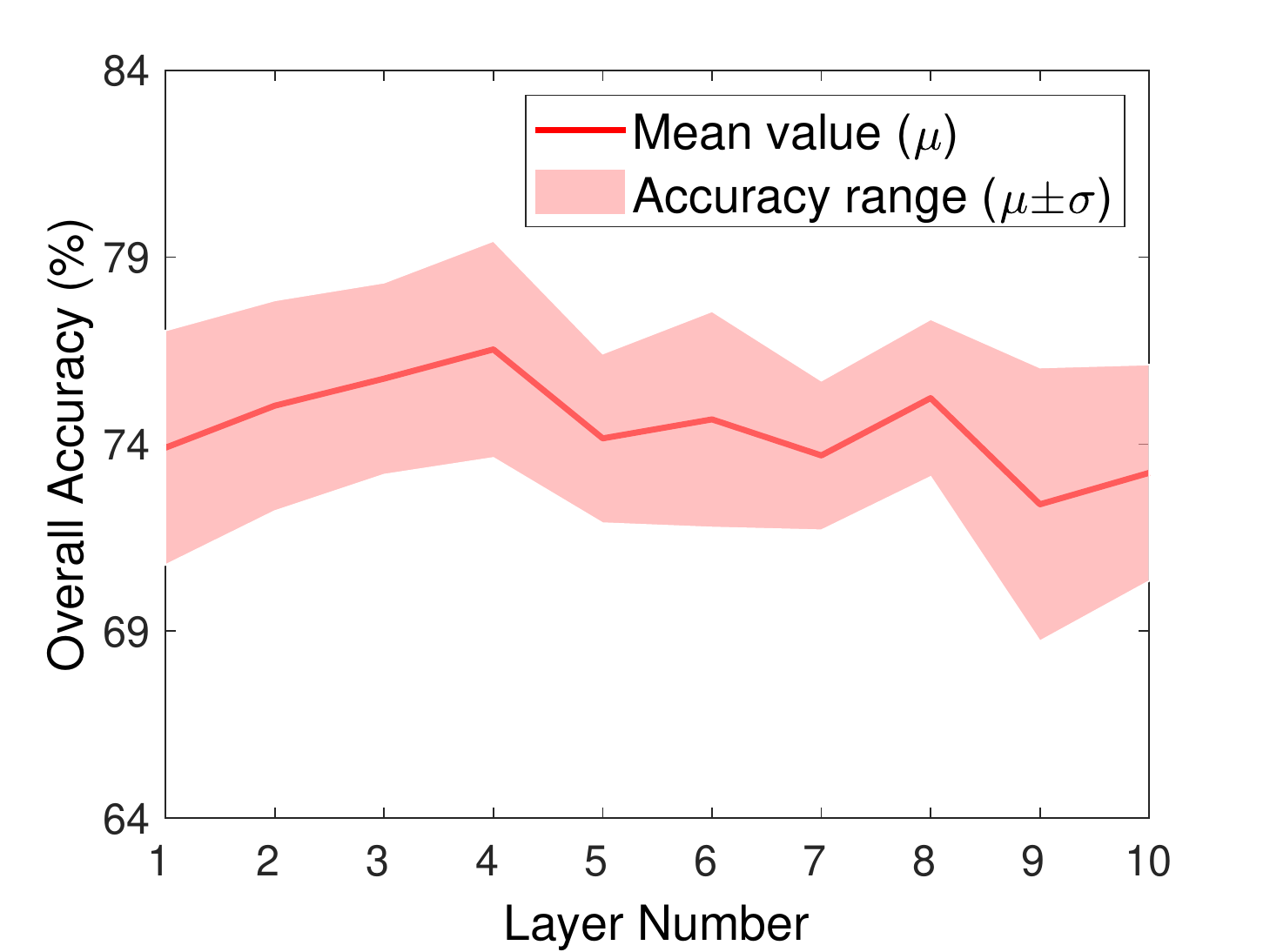}}
  \subfigure[]{\includegraphics[width=0.49\linewidth]{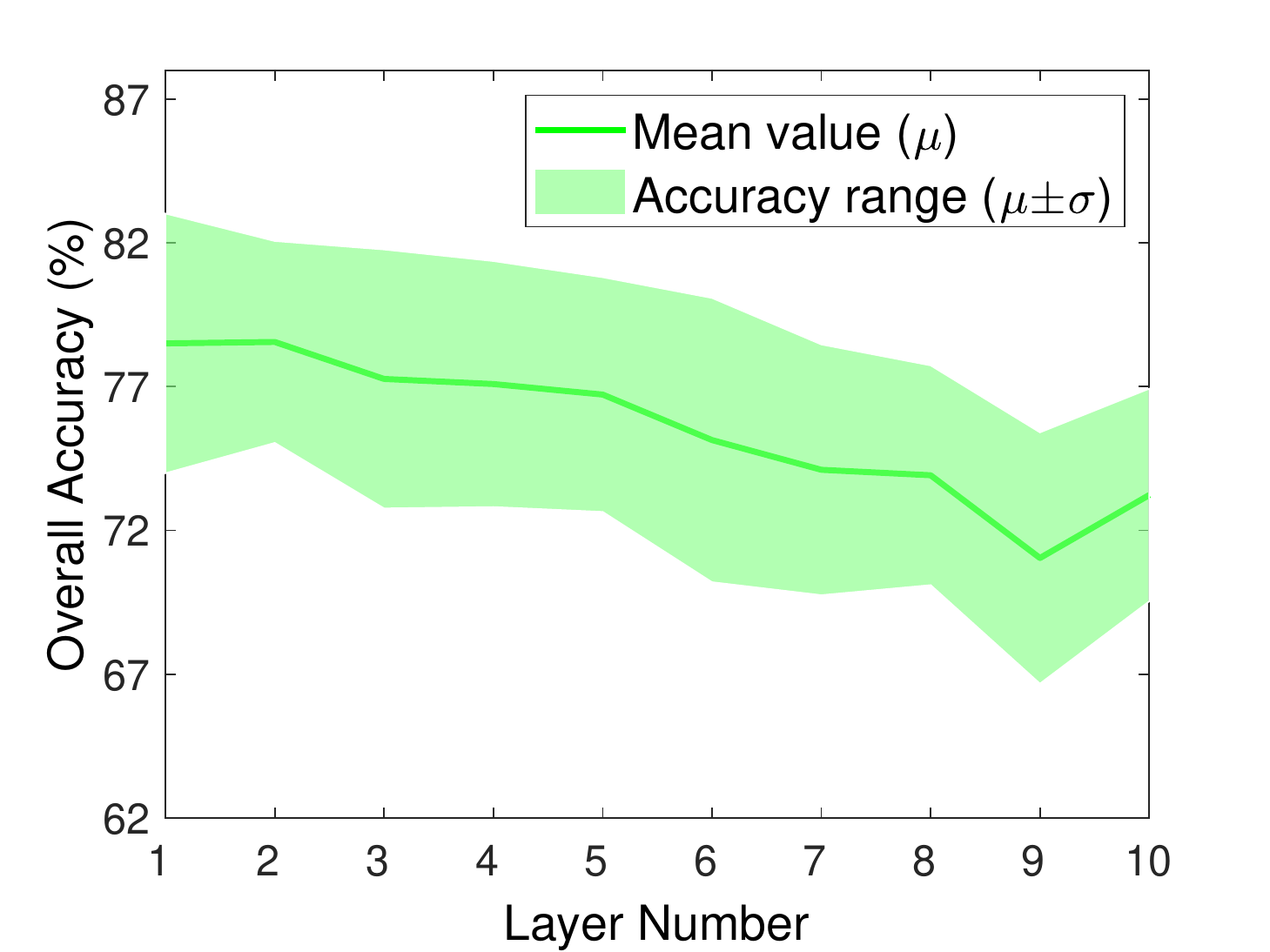}}\\
  \subfigure[]{\includegraphics[width=0.49\linewidth]{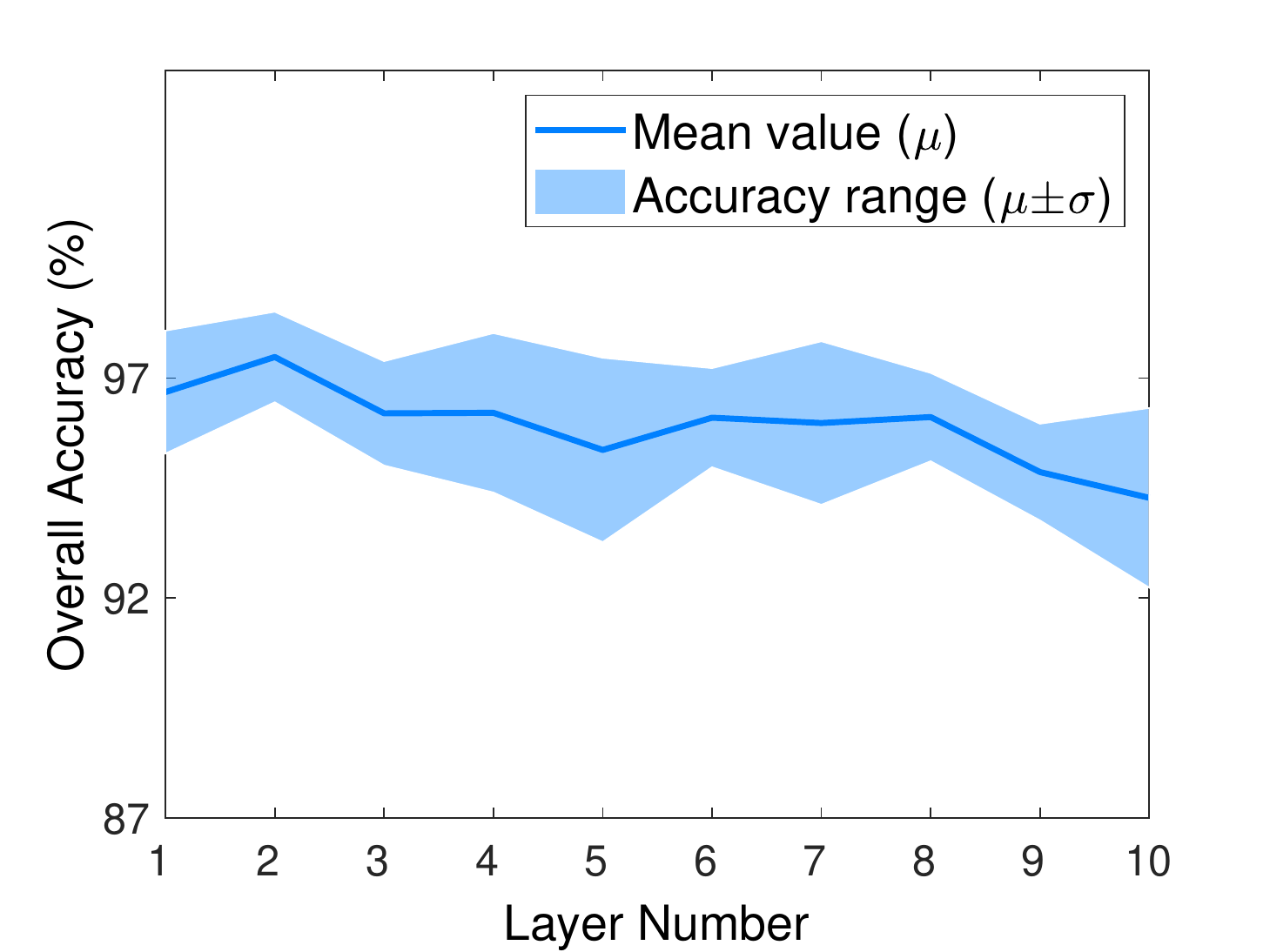}}
  \subfigure[]{\includegraphics[width=0.49\linewidth]{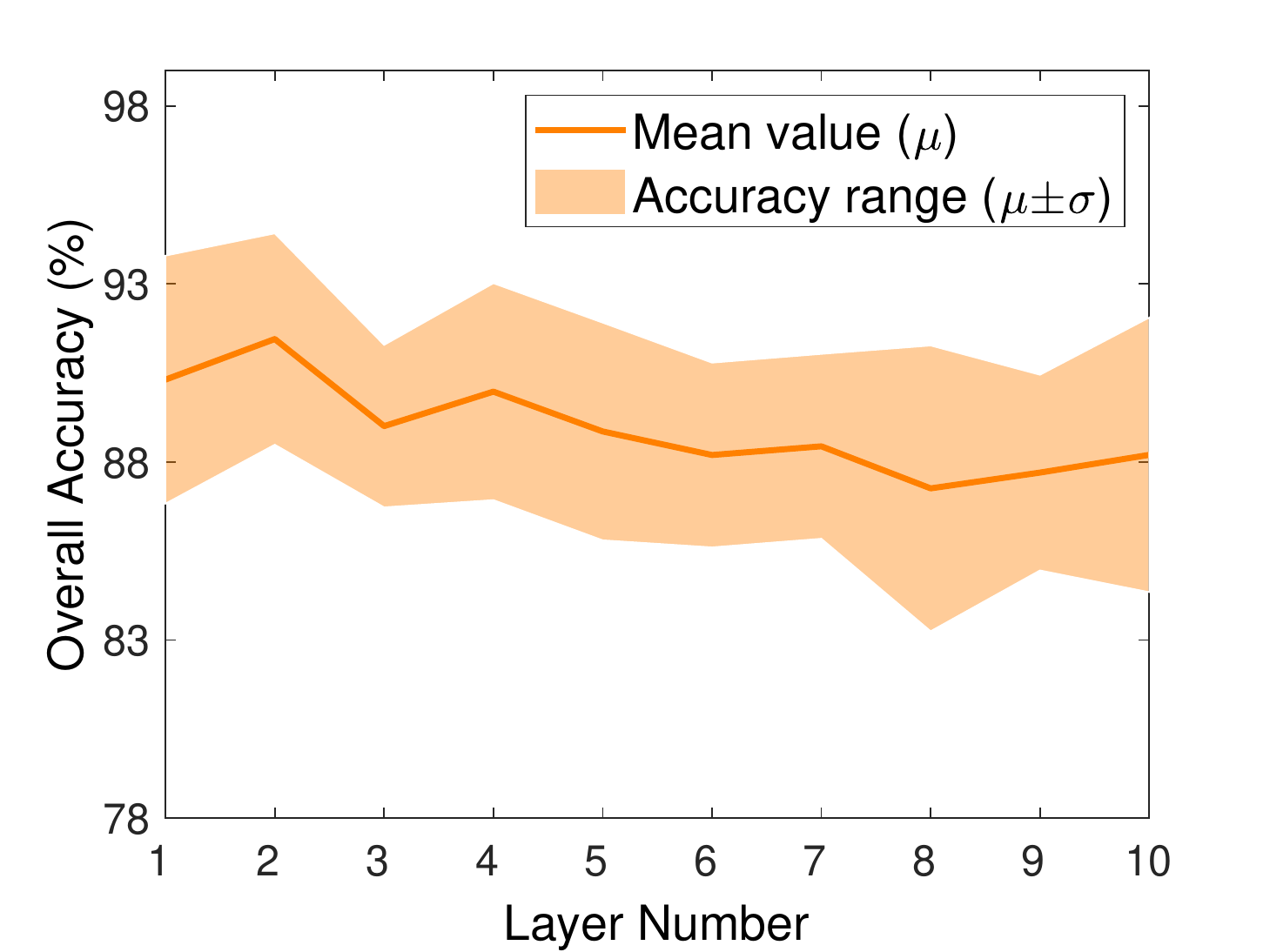}}
  \caption{Relationships between the layer number and the overall accuracy for 3-D HK-CLS on different datasets when block number equals 3 . (a) Indian Pines. (b) Pavia University. (c) Kennedy Space Center. (d) Salinas Valley.}
  \label{layer_acc}
\end{figure}

\begin{table}[t]
  \caption{The determined hyper parameters including block number ($M$) and layer number ($N$) on different datasets}
  \newcommand{\tabincell}[2]{\begin{tabular}{@{}#1@{}}#2\end{tabular}}
  \centering
  \label{scheme_222_842}
  \resizebox{0.8\linewidth}{!}{
  \begin{tabular}{|l|cc|cc|}
  \hline
  \multirow{2}{*}{Hyper Parameters}& \multicolumn{2}{c|}{Indian Pines} & \multicolumn{2}{c|}{Pavia University}  \\
  \cline{2-5}
  & $M$ &$N$& $M$ &$N$\\
  \hline
  1-D HK-CLS              & 6 & 5 &4 &1  \\
  3-D HK-CLS               &3 & 4 &3 &2 \\
  3-D HK-SEG &3 & 1 &3 &1  \\
  \hline
  \multirow{2}{*}{Hyper Parameters}& \multicolumn{2}{c|}{Kennedy Space Center} & \multicolumn{2}{c|}{Salinas Valley}  \\
  \cline{2-5}
  & $M$ &$N$ & $M$ &$N$\\
  \hline
  1-D HK-CLS              &3& 2 & 4 & 1 \\
  3-D HK-CLS               & 3 &2  &3 &2  \\
  3-D HK-SEG &3 & 1 &3 &1   \\
  \hline
\end{tabular}
  }
  \label{block_layer_set}
\end{table}

In the proposed methods, the number of blocks $M$ and the number of layers $N$ inside one block of the search space framework are the only hyperparameters that need to be manually configured. To obtain better architectures used for HSI classification, with the validation set, we construct the relationships between them and the OA of the searched network structures.

For the network of 1-D HK-CLS, the relationships between the above hyperparameters and accuracies are shown in Figure \ref{block_layer_acc}, it is interesting to see that, in most scenarios, model performances present a trend that it is decreasing with the complexity of the network, and the lighter model performs better. This may be because more trainable parameters are carried in larger networks, causing the phenomenon of overfitting and affecting the accuracies. While in the Indian Pines dataset, accuracies first increase and then decrease with the change of these two hyperparameters. When the block number is set to 6 and the layer number equals 5, the proposed method performs the best.

The largest block number of 3-D HK-CLS only can be set to 3 with the consideration of the size of input patches. Therefore, only the relationships between layer numbers and accuracies need to be concerned, and the results are depicted in Figure \ref{layer_acc}. Besides the OAs are changed, we can observe that upper bounds $\mu+\sigma$, lower bounds $\mu-\sigma$, and average values $\mu$ present similar profiles, indicating that the accuracies possess roughly the same variation ranges $\sigma$ when using different layer numbers. Thus, the most suitable layer number can be directly determined through the value of $\mu$. Using similar approaches, we also find the optimal layer number of 3-D HK-SEG.

The final block number and layer number of the above three kinds of networks are determined in the light of accuracies. For example, the block number and the layer number of 1-D HK-CLS are separately decided to be 6 and 5 on the Indian Pines dataset, and all the results have been listed in Table \ref{block_layer_set}, which will be located as the basic configurations in later experiments. Note that the above results of 3-D HK-CLS and 3-D HK-SEG are derived by following the settings in \cite{3dcnn,SSRN}. Therefore, the 3-D convolution is successively replaced by 1-D convolution and 2-D depth-wise convolution to improve network flexibility, meaning spectral features are first extracted, and then spatial contexts are aggregated. The subsequent studies will determine the most suitable way of 3-D convolution decomposition for each kind of network.

\subsection{Ablation Study}

\begin{figure}[t]
  \centering
  \subfigure[]{\includegraphics[width=0.49\linewidth]{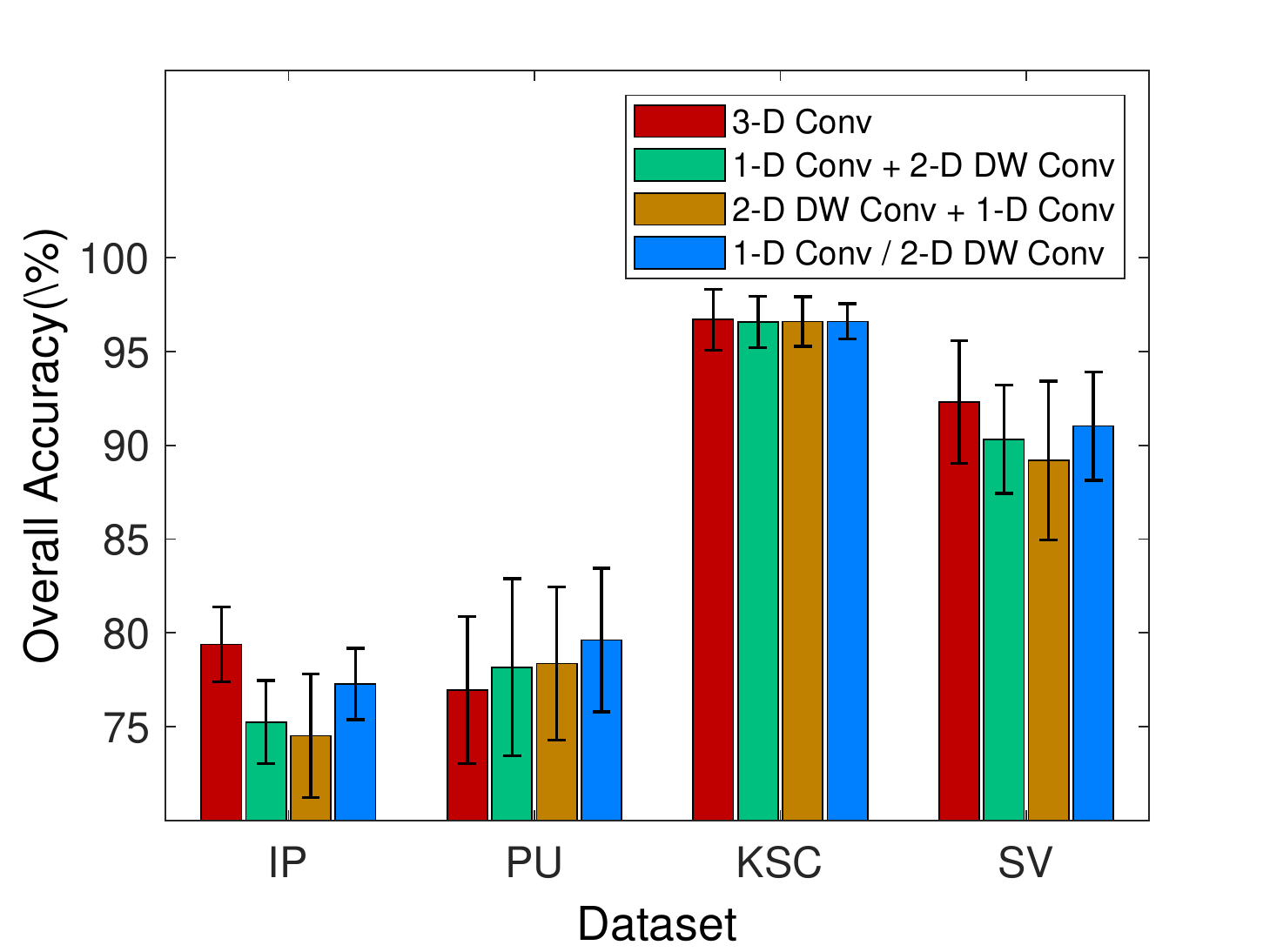}}
  \subfigure[]{\includegraphics[width=0.49\linewidth]{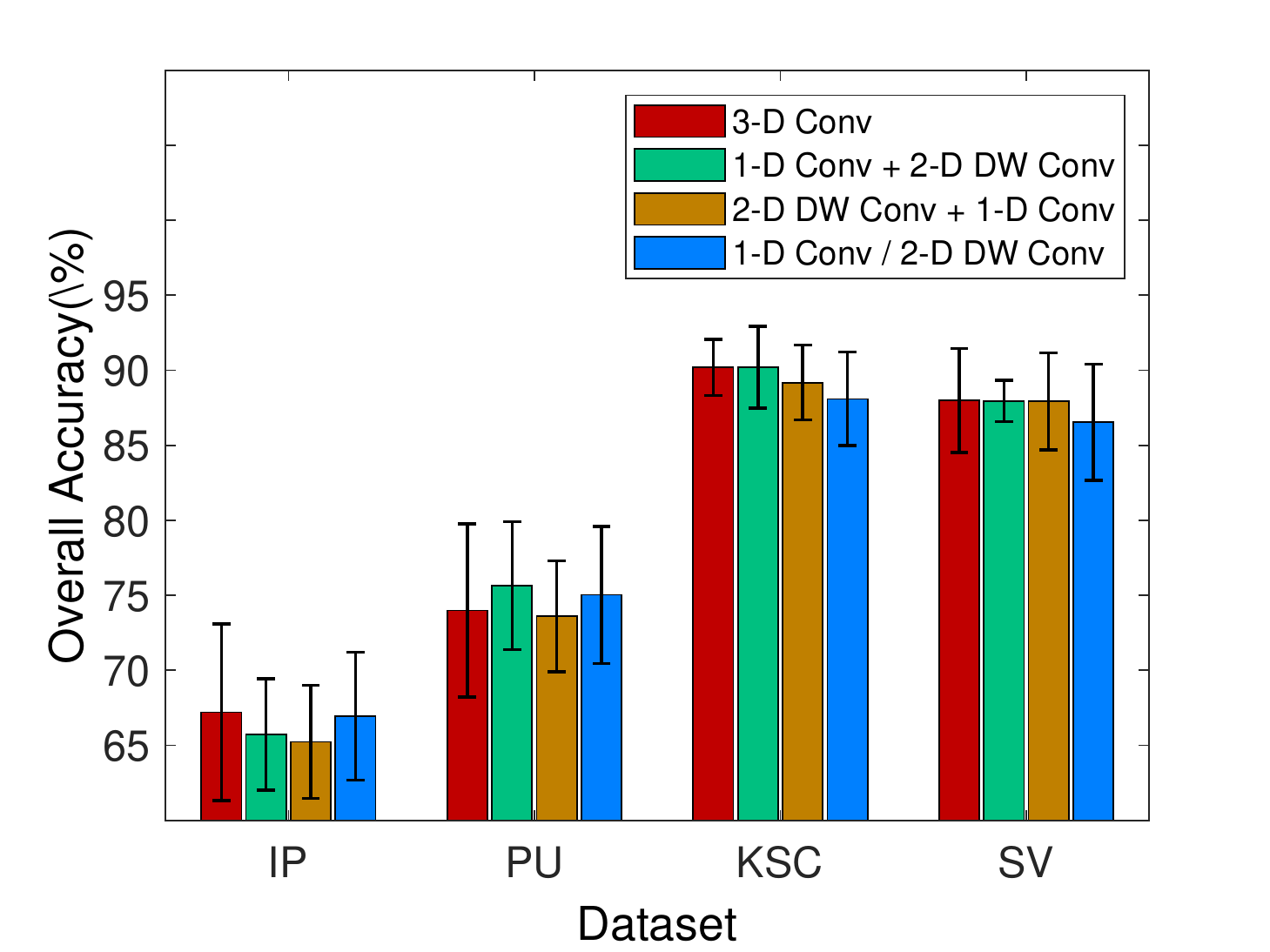}}\\
  \caption{Performance of the networks using the 3-D convolution in four different forms. (a) 3-D HK-CLS. (b) 3-D HK-SEG. Here, IP: Indian Pines. PU: Pavia University. KSC: Kennedy Space Center and SV: Salinas Valley.}
  \label{3dconv_ablation}
\end{figure}

\begin{table}[t]
  \caption{The determined 3-D convolution settings of 3-D HK-CLS and 3-D HK-SEG on different datasets}
  \newcommand{\tabincell}[2]{\begin{tabular}{@{}#1@{}}#2\end{tabular}}
  \centering
  \label{scheme_222_842}
  \resizebox{\linewidth}{!}{
  \begin{tabular}{|l|l|l|}
  \hline
  3-D Convolution & Indian Pines & Pavia University  \\
  \hline
  3-D HK-CLS               &3-D Conv &  1-D Conv / 2-D DW Conv \\
  3-D HK-SEG &  3-D Conv & 1-D Conv + 2-D DW Conv  \\
  \hline
  3-D Convolution & Kennedy Space Center & Salinas Valley  \\
  \hline
  3-D HK-CLS               & 3-D Conv & 3-D Conv  \\
  3-D HK-SEG & 1-D Conv + 2-D DW Conv & 3-D Conv   \\
  \hline
\end{tabular}
  }
  \label{3dconv_set}
\end{table}

\begin{table}[t]
  \caption{Accuracy comparisons of HKNAS with one and two-tier optimizations (\%), ``T'' means adopting the two-tier optimization.}
  \newcommand{\tabincell}[2]{\begin{tabular}{@{}#1@{}}#2\end{tabular}}
  \centering
  \resizebox{\linewidth}{!}{
  \begin{tabular}{|l|cccccc|}
  \hline
  Method & \tabincell{c}{1-D\\HK-CLS (T)} & \tabincell{c}{1-D\\HK-CLS} & \tabincell{c}{3-D\\HK-CLS (T)} & \tabincell{c}{3-D\\HK-CLS} & \tabincell{c}{3-D\\HK-SEG (T)} & \tabincell{c}{3-D\\HK-SEG}  \\
  \hline
  \bfseries Indian Pines & \tabincell{c}{54.79 \\ $\pm$2.88} & \bfseries \tabincell{c}{56.19 \\ $\pm$2.04} & \tabincell{c}{76.34 \\$\pm$2.51} & \bfseries\tabincell{c}{79.38 \\$\pm$2.00} & \tabincell{c}{66.96 \\$\pm$5.35} & \bfseries\tabincell{c}{67.21 \\$\pm$5.88}  \\
  \hline\hline
  \bfseries Pavia University & \bfseries \tabincell{c}{68.64 \\ $\pm$3.34} & \tabincell{c}{68.39 \\ $\pm$3.91} & \tabincell{c}{77.62 \\ $\pm$5.68} & \bfseries\tabincell{c}{79.62 \\ $\pm$3.83} & \tabincell{c}{75.33 \\$\pm$5.06} & \bfseries\tabincell{c}{75.64 \\$\pm$4.26}  \\
  \hline\hline
  \bfseries Kennedy Space Center & \tabincell{c}{85.76 \\ $\pm$0.48} & \bfseries\tabincell{c}{85.97 \\ $\pm$0.78} & \bfseries\tabincell{c}{97.09 \\ $\pm$1.71} & \tabincell{c}{96.70 \\ $\pm$1.62} & \tabincell{c}{89.92 \\ $\pm$2.54} & \bfseries\tabincell{c}{90.21 \\ $\pm$2.73} \\
  \hline\hline
  \bfseries Salinas Valley &\tabincell{c}{84.01 \\ $\pm$2.08} & \bfseries\tabincell{c}{85.00 \\ $\pm$2.43} & \tabincell{c}{91.48 \\ $\pm$3.21} & \bfseries\tabincell{c}{92.31 \\ $\pm$3.28} & \tabincell{c}{87.95 \\ $\pm$2.30} & \bfseries\tabincell{c}{87.99 \\ $\pm$3.46} \\
  \hline
\end{tabular}
  }
  \label{one_two_tier}
\end{table}

\begin{table*}[t]
  \caption{Accuracy comparisons of different methods on four classical datasets (\%)}
  \newcommand{\tabincell}[2]{\begin{tabular}{@{}#1@{}}#2\end{tabular}}
  \centering
  \label{scheme_222_842}
  \resizebox{\linewidth}{!}{
  \begin{tabular}{|l|ccccc|cccccc|cc|}
  \hline
   Group & \multicolumn{5}{c|}{1-D Pixel-level} & \multicolumn{6}{c|}{3-D Pixel-level}  & \multicolumn{2}{c|}{3-D Image-level}  \\
   \hline\hline
  \bfseries Indian Pines & RBF-SVM & 1-D CNN & 1-D Auto-CNN & 1-D P-DARTS & 1-D HK-CLS & 3-D CNN & 3-D Auto-CNN & 3-D P-DARTS & NetworkAdjustment & SSTN & 3-D HK-CLS & 3-D ANAS & 3-D HK-SEG  \\
  \hline
  OA & \bfseries \tabincell{c}{56.78 \\ $\pm$2.02} & \tabincell{c}{53.87 \\ $\pm$3.93} & \tabincell{c}{52.54 \\ $\pm$3.62}&\tabincell{c}{50.98 \\ $\pm$2.49} & \tabincell{c}{56.19 \\ $\pm$2.04} & \tabincell{c}{63.25 \\ $\pm$2.54} & \tabincell{c}{72.64 \\ $\pm$3.80} & \tabincell{c}{74.10 \\$\pm$1.66} & \tabincell{c}{73.75\\$\pm$3.70} &\tabincell{c}{51.18\\$\pm$3.93} & \bfseries\tabincell{c}{79.38 \\$\pm$2.00} & \tabincell{c}{59.51 \\ $\pm$4.78} & \bfseries\tabincell{c}{67.21 \\$\pm$5.88}  \\
  AA & \tabincell{c}{68.03 \\$\pm$0.60} & \tabincell{c}{66.51 \\ $\pm$2.19} & \tabincell{c}{64.51 \\ $\pm$1.49}& \tabincell{c}{62.76 \\$\pm$1.69}& \bfseries\tabincell{c}{68.65 \\ $\pm$1.80} & \tabincell{c}{77.60 \\ $\pm$1.65} & \tabincell{c}{83.74 \\$\pm$2.09} & \tabincell{c}{83.21\\$\pm$1.94} & \tabincell{c}{84.17\\$\pm$3.92} & \tabincell{c}{49.97\\$\pm$5.20} & \bfseries\tabincell{c}{87.60 \\ $\pm$1.37} & \tabincell{c}{75.02 \\ $\pm$2.63} & \bfseries\tabincell{c}{79.00 \\ $\pm$4.49}  \\
  Kappa & \bfseries\tabincell{c}{51.58 \\$\pm$2.06} & \tabincell{c}{48.51 \\ $\pm$3.82} & \tabincell{c}{47.10 \\ $\pm$3.61}&  \tabincell{c}{45.20 \\$\pm$2.76} & \tabincell{c}{51.18 \\ $\pm$2.22} & \tabincell{c}{58.91 \\ $\pm$2.75} & \tabincell{c}{69.30 \\ $\pm$4.09} & \tabincell{c}{70.88\\$\pm$1.82} & \tabincell{c}{70.51\\$\pm$4.05} & \tabincell{c}{44.47\\$\pm$3.72} & \bfseries\tabincell{c}{76.74 \\ $\pm$2.20} & \tabincell{c}{54.90 \\$\pm$5.02} & \bfseries\tabincell{c}{62.95 \\$\pm$6.67}  \\
  \hline\hline
  \bfseries Pavia University & RBF-SVM & 1-D CNN & 1-D Auto-CNN & 1-D P-DARTS & 1-D HK-CLS & 3-D CNN & 3-D Auto-CNN & 3-D P-DARTS & NetworkAdjustment & SSTN &  3-D HK-CLS & 3-D ANAS & 3-D HK-SEG  \\
  \hline
  OA & \tabincell{c}{65.29 \\ $\pm$3.71} & \tabincell{c}{65.44 \\$\pm$4.19} & \bfseries\tabincell{c}{70.21 \\$\pm$3.87} & \tabincell{c}{69.81 \\$\pm$3.09} & \tabincell{c}{68.39 \\ $\pm$3.91} & \tabincell{c}{70.82 \\ $\pm$2.76} & \tabincell{c}{73.85 \\ $\pm$5.81} & \tabincell{c}{78.91\\$\pm$3.80} & \tabincell{c}{72.05\\$\pm$4.22}& \tabincell{c}{78.26\\$\pm$5.21} & \bfseries\tabincell{c}{79.62 \\ $\pm$3.83} & \tabincell{c}{73.22 \\$\pm$2.57} & \bfseries\tabincell{c}{75.64 \\$\pm$4.26}  \\
  AA & \tabincell{c}{76.90 \\$\pm$1.74} & \tabincell{c}{75.83 \\ $\pm$3.52} & \tabincell{c}{77.78 \\ $\pm$1.64}& \tabincell{c}{77.76 \\$\pm$2.50}& \bfseries\tabincell{c}{78.81 \\ $\pm$1.71} & \tabincell{c}{71.55 \\ $\pm$1.96} & \tabincell{c}{78.78 \\$\pm$2.96} & \tabincell{c}{81.06\\$\pm$2.55} &\tabincell{c}{72.87\\$\pm$1.83}& \tabincell{c}{81.21\\$\pm$5.02} & \bfseries\tabincell{c}{82.46 \\$\pm$3.14} & \tabincell{c}{79.31 \\ $\pm$2.26} & \bfseries\tabincell{c}{79.84 \\$\pm$2.20}  \\
  Kappa & \tabincell{c}{57.30 \\$\pm$3.79} & \tabincell{c}{57.33 \\ $\pm$4.85} & \bfseries\tabincell{c}{62.49 \\$\pm$4.21} & \tabincell{c}{62.06 \\$\pm$3.41} & \tabincell{c}{60.76 \\ $\pm$4.08} & \tabincell{c}{62.44 \\ $\pm$2.93} & \tabincell{c}{67.52 \\ $\pm$6.41} & \tabincell{c}{73.34\\$\pm$4.34} & \tabincell{c}{64.70\\$\pm$4.40} & \tabincell{c}{71.69\\$\pm$6.72} & \bfseries\tabincell{c}{74.21 \\$\pm$4.57} & \tabincell{c}{65.97 \\ $\pm$2.63} & \bfseries\tabincell{c}{69.23 \\$\pm$4.58}  \\
  \hline\hline
  \bfseries Kennedy Space Center & RBF-SVM & 1-D CNN & 1-D Auto-CNN & 1-D P-DARTS  & 1-D HK-CLS & 3-D CNN & 3-D Auto-CNN & 3-D P-DARTS & NetworkAdjustment & SSTN & 3-D HK-CLS & 3-D ANAS & 3-D HK-SEG  \\
  \hline
  OA & \tabincell{c}{84.14 \\ $\pm$1.48} & \tabincell{c}{75.50 \\ $\pm$2.22} & \tabincell{c}{85.74 \\ $\pm$0.75} &\tabincell{c}{85.48 \\ $\pm$0.96} & \bfseries\tabincell{c}{85.97 \\ $\pm$0.78} & \tabincell{c}{88.85 \\ $\pm$2.08} & \tabincell{c}{95.69 \\ $\pm$1.00} &\tabincell{c}{96.34\\$\pm$0.88} & \tabincell{c}{89.17\\$\pm$3.68} & \tabincell{c}{94.83\\$\pm$1.06} & \bfseries\tabincell{c}{96.70 \\ $\pm$1.62} & \tabincell{c}{87.58 \\ $\pm$0.53} & \bfseries\tabincell{c}{90.21 \\ $\pm$2.73} \\
  AA & \tabincell{c}{80.31 \\ $\pm$1.03} & \tabincell{c}{70.03 \\ $\pm$2.98} & \tabincell{c}{81.30 \\ $\pm$0.86} & \tabincell{c}{81.52 \\$\pm$0.97} & \bfseries\tabincell{c}{81.68 \\ $\pm$1.08} & \tabincell{c}{87.73 \\ $\pm$2.44} & \tabincell{c}{95.32 \\ $\pm$0.96} & \tabincell{c}{95.68\\$\pm$0.77} & \tabincell{c}{87.69\\$\pm$3.33} & \tabincell{c}{92.09\\$\pm$1.56} & \bfseries\tabincell{c}{96.17 \\ $\pm$1.70} & \tabincell{c}{84.90 \\ $\pm$0.38} & \bfseries\tabincell{c}{87.43 \\ $\pm$3.12}  \\
  Kappa & \tabincell{c}{82.37 \\ $\pm$1.62} & \tabincell{c}{72.79 \\ $\pm$2.41} & \tabincell{c}{84.11 \\ $\pm$0.82}& \tabincell{c}{83.83 \\$\pm$1.06} & \bfseries\tabincell{c}{84.36 \\ $\pm$0.87} & \tabincell{c}{87.60 \\ $\pm$2.31} & \tabincell{c}{95.19 \\ $\pm$1.11} & \tabincell{c}{95.92\\$\pm$0.98} & \tabincell{c}{87.93\\$\pm$4.10} & \tabincell{c}{94.23\\$\pm$1.18} & \bfseries\tabincell{c}{96.32 \\ $\pm$1.81} & \tabincell{c}{86.15 \\ $\pm$0.58} & \bfseries\tabincell{c}{89.08 \\ $\pm$3.03}  \\
  \hline\hline
  \bfseries Salinas Valley & RBF-SVM & 1-D CNN & 1-D Auto-CNN & 1-D P-DARTS & 1-D HK-CLS & 3-D CNN & 3-D Auto-CNN & 3-D P-DARTS & NetworkAdjustment & SSTN & 3-D HK-CLS & 3-D ANAS & 3-D HK-SEG  \\
  \hline
  OA & \tabincell{c}{82.88 \\ $\pm$1.32} & \tabincell{c}{77.13 \\ $\pm$1.58} & \tabincell{c}{84.29 \\ $\pm$2.06}& \tabincell{c}{84.87\\ $\pm$2.27} &\bfseries\tabincell{c}{85.00 \\ $\pm$2.43} & \tabincell{c}{84.88 \\ $\pm$3.80} & \tabincell{c}{88.24 \\ $\pm$2.96} & \tabincell{c}{92.04\\$\pm$2.84} & \tabincell{c}{83.99\\$\pm$2.28} & \tabincell{c}{84.66\\$\pm$4.43} & \bfseries\tabincell{c}{92.31 \\ $\pm$3.28} & \tabincell{c}{83.74 \\ $\pm$0.17} & \bfseries\tabincell{c}{87.99 \\ $\pm$3.46} \\
  AA & \tabincell{c}{90.28 \\ $\pm$0.79} & \tabincell{c}{84.43 \\ $\pm$1.75} & \tabincell{c}{91.04 \\ $\pm$1.68}& \tabincell{c}{91.66 \\$\pm$1.24} & \bfseries\tabincell{c}{91.78 \\ $\pm$1.26} & \tabincell{c}{89.90 \\ $\pm$2.32} & \tabincell{c}{91.96 \\ $\pm$2.49} & \tabincell{c}{95.37\\$\pm$1.81} & \tabincell{c}{89.32\\$\pm$1.23} & \tabincell{c}{91.43\\$\pm$6.54} & \bfseries\tabincell{c}{96.24 \\ $\pm$1.54} & \tabincell{c}{89.46 \\$\pm$0.71} & \bfseries\tabincell{c}{93.27 \\ $\pm$2.03}  \\
  Kappa & \tabincell{c}{81.01 \\ $\pm$1.45} & \tabincell{c}{74.69 \\ $\pm$1.75} & \tabincell{c}{82.57 \\ $\pm$2.26} & \tabincell{c}{83.01 \\$\pm$2.47} & \bfseries\tabincell{c}{83.36 \\ $\pm$2.65} & \tabincell{c}{83.26 \\ $\pm$4.19} & \tabincell{c}{86.97 \\ $\pm$3.25} & \tabincell{c}{91.16\\$\pm$3.16} & \tabincell{c}{82.22\\$\pm$2.43} & \tabincell{c}{82.78\\$\pm$4.98} & \bfseries\tabincell{c}{91.48 \\ $\pm$3.61} & \tabincell{c}{81.90 \\ $\pm$0.20} & \bfseries\tabincell{c}{86.68 \\ $\pm$3.82} \\
  \hline
\end{tabular}
  }
  \label{classical_acc_compare}
\end{table*}

\noindent\textbf{3-D Convolution.} In order to obtain optimal network structures, besides the above hyperparameters, we also determine the specific form for 3-D convolutions in the HKNAS part of 3-D HK-CLS and 3-D HK-SEG. In Section III-A, we have proved that a standard 3-D convolution can be substituted by a 1-D convolution and a 2-D depth-wise convolution. If the 1-D convolution comes first and the 2-D depth-wise convolution is in back, as the formula (\ref{1d_2ddw}) shown, we call this form ``1-D Conv + 2-D DW Conv'', where ``+'' represents serial. Analogously, the order of 1-D convolution and 2-D depth-wise convolution can be exchanged, generating the form of ``2-D DW Conv + 1-D Conv''. Furthermore, we also evaluate a form where two parallel branches are presented to separately extract spectral and spatial features using 1-D convolution and 2-D depth-wise convolution, respectively, and then the obtained spectral and spatial features are added together. This form is named ``1-D Conv / 2-D DW Conv''. Here, ``/'' symbolizes parallel.

Figure \ref{3dconv_ablation} displays the performances of 3-D HK-CLS and 3-D HK-SEG using the above four 3-D convolution forms on different datasets. It can be seen that different 3-D convolution forms may have different impacts on network performances in some scenarios, such as on the Indian Pines dataset when adopting 3-D HK-CLS, and there are the most suitable 3-D convolution forms for different datasets. For convenience, we directly choose the 3-D convolution form that possesses the highest accuracy. For example, the optimal 3-D convolution structure of the Indian Pines dataset when using 3-D HK-CLS is standard 3-D convolution. The final selections are shown in Table \ref{3dconv_set}, where ``3-D Conv’’ means standard 3-D convolution, and these results will be served as the basic structure settings for the spectral-spatial feature classification networks in later performance comparison experiments.

\noindent\textbf{One-tier Optimization.} We compare different HKNAS networks that separately adopt one-tier and two-tier optimizations. In the two-tier group, following DARTS \cite{darts}, structural parameters are independently defined and are alternately optimized with kernel weights by separately employing half of the training set (introduced in the last paragraph of Section IV-A). The other settings, such as hyperparameters, are remaining the same. The comparison results have been listed in Table \ref{one_two_tier}. It can be seen that the networks employing one-tier optimization perform better than the two-tier counterparts in the majority of cases since relationships between architectures and network weights are strengthened, verifying the effectiveness of the proposed methods.

\subsection{Performance Comparison}

We implement the quantitative performance comparisons of the proposed methods with the state-of-the-art approaches including RBF-SVM, 1-DCNN, 1-D Auto-CNN, 1-D P-DARTS, 3-D CNN, 3-D Auto-CNN, 3-D P-DARTS, NetworkAdjustment, SSTN, and 3-D ANAS. Before presenting the comparisons, we conduct brief introductions of these methods.

\begin{itemize}
  \item [1)] RBF-SVM: The raw spectral vectors are directly used for pixel-level classification. In our implementation, two important hyperparameters $C$ and $\gamma$ are determined by fivefold cross-validation searching inside a grid in the range of $\{2^{5}, 2^{-4}, \cdots, 2^{19}\}$ and $\{2^{-15}, 2^{-14}, \cdots, 2^{5}\}$ through following \cite{3dcnn}. ``RBF'' means the radial basis function kernel. This algorithm is implemented with LIBSVM library \cite{libsvm}.
  \item [2)] 1-D CNN: Pixel-level classification by 1-D deep networks, where spectral vectors are fed into a network that includes 1-D convolutional and pooling layers. The related configurations such as depth or kernel sizes follow \cite{3dcnn}.
  \item [3)] 1-D Auto-CNN: Pixel-level classification by 1-D deep networks, where three cells are presented. The architecture inside each cell is searched by DARTS \cite{darts}. The candidate operations in the search space are all 1-D convolutions or poolings, such as the 1-D separable convolution with various kernel sizes.
  \item [4)] 1-D P-DARTS: Pixel-level classification by 1-D deep networks. In addition to the DARTS of 1-D Auto-CNN is replaced by P-DARTS \cite{pdarts}, other settings are the same as 1-D Auto-CNN.
  \item [5)] 3-D CNN: Pixel-level classification by 3-D deep networks, where spectral-spatial cubes are fed into a network that includes standard 3-D convolutional and pooling layers. The related configurations such as depth or kernel sizes follow \cite{3dcnn}.
  \item [6)] 3-D Auto-CNN: Pixel-level classification by 3-D deep networks, where three cells are presented. The architecture inside each cell is searched by DARTS. In their implementation, the candidate operations in the search space are 2-D convolutions or poolings, such as the 2-D separable convolution or 2-D dilated convolution with various kernel sizes.
  \item [7)] 3-D P-DARTS: Pixel-level classification by 3-D deep networks. In addition to the DARTS of 3-D Auto-CNN is replaced by P-DARTS, other settings are the same as 3-D Auto-CNN.
  \item [8)] NetworkAdjustment: Pixel-level classification by 3-D deep networks, which are obtained by changing channel and block numbers inside stages of ResNet-20 with the consideration of resource utilization, such as the floating-point operations per second (FLOPs). The sizes of input patches are the same as 3-D Auto-CNN and 3-D P-DARTS.
  \item [9)] SSTN: Pixel-level classification by 3-D deep networks, whose architectures are searched by separately implementing the bi-level optimization in two factorized subspaces that in layer-level and block-level, respectively.
  \item [10)] 3-D ANAS: Image-level classification by 3-D deep networks, where whole images are fed into a network that includes parallel multiple cells possessing different channel numbers. The architecture inside each cell is searched by DARTS, while the optimal paths between different cells are determined with the viterbi algorithm.
 \end{itemize}

 For the convenience of comparisons, according to the type of input and output, we divide the above methods into three groups: 1-D pixel-level, 3-D pixel-level, and 3-D image-level. Here, the RBF-SVM, 1-DCNN, 1-D Auto-CNN, 1-D P-DARTS, and our 1-D HK-CLS belong to the 1-D pixel-level group, the 3-D CNN, 3-D Auto-CNN, 3-D P-DARTS, NetworkAdjustment, SSTN and the proposed 3-D HK-CLS are included in the 3-D pixel-level group, while the 3-D image-level group contains 3-D ANAS and 3-D HK-SEG.

 \subsubsection{Accuracy on Classical Datasets}

 We first evaluate the above methods on four classical datasets and the accuracies are listed in Table \ref{classical_acc_compare}. Here, we record the mean value and standard deviation of 10 times trials, and the best performances in each group are marked in bold. It can be seen that the methods in the 3-D pixel-level group perform better than the 1-D pixel-level group since the surroundings of the target pixel are considered, meaning besides spectral information, spatial features are also extracted for classification. Although image-level methods have been shown can achieve higher accuracies than pixel-level methods \cite{fcontnet,freenet,ssfcn,enl_fcn}. In this paper, the performances of the 3-D image-level group do not surpass the 3-D pixel-level group because the number of training samples is very small (10 samples per class). In the case of few samples, the image-level methods receiving the whole image are trained with a label map where the information of very few locations is knowable, bringing about difficulties that global contexts can not be effectively captured. It seems that these methods are forcibly memorizing the labels of these locations, causing a risk of overfitting and decreasing the generalization ability. While the stochastic mini-batch training ensures pixel-level methods can effectively recognize the inherent characteristics of each category, so as to naturally identify new samples. In addition, the upsampling operations such as the bilinear interpolation may introduce the information of unknown locations into existing labeled features, and errors are inevitably involved.

In the 1-D pixel-level group, as the representation of traditional methods, RBF-SVM achieves competitive results, and even obtains the best OA on the Indian Pines dataset, showing the potential of conventional methods. Compared with the handcrafted 1-D CNN, 1-D Auto-CNN, 1-D P-DARTS and our 1-D HK-CLS perform better since network structures are automatically found by computer, avoiding the interferences of human factors that are usually introduced in manual network construction procedures. In NAS-based classification methods, benefitting from the HKNAS that employs hyper kernels to directly generate structural parameters, constructing the connections between architectures and network weights, and transforming the difficult dual optimization problem into a simple one-tier optimization problem. Our 1-D HK-CLS outperforms 1-D Auto-CNN and 1-D P-DARTS, especially in Kennedy Space Center and Salinas Valley datasets. In the groups of 3-D pixel-level and 3-D image-level, besides the above merits, taking the advantages of diverse 3-D convolution forms, 3-D HK-CLS and 3-D HK-SEG outperform other methods and achieve the best, since the flexibilities of network structures are further improved. We notice an advanced 3-D image-level classification method, 3-D ANAS, is seriously affected by the few sample situation, and its accuracies on Indian Pines and Salinas Valley are even worse than 1-D pixel-level methods. As it has been discussed, image-level classification methods have higher overfitting risks than others. Nevertheless, our 3-D HK-SEG still performs well.

 \begin{figure}[t]
  \centering
  \subfigure[]{\includegraphics[width=0.15\linewidth]{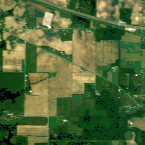}}
  \subfigure[]{\includegraphics[width=0.15\linewidth]{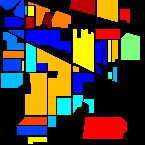}}
  \subfigure[]{\includegraphics[width=0.15\linewidth]{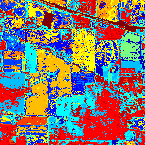}}
  \subfigure[]{\includegraphics[width=0.15\linewidth]{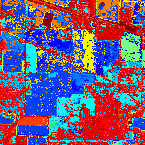}}
  \subfigure[]{\includegraphics[width=0.15\linewidth]{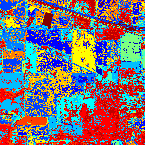}}\\
  \subfigure[]{\includegraphics[width=0.15\linewidth]{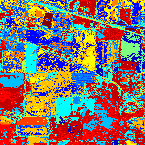}}
  \subfigure[]{\includegraphics[width=0.15\linewidth]{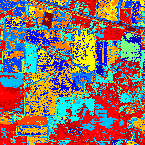}}
  \subfigure[]{\includegraphics[width=0.15\linewidth]{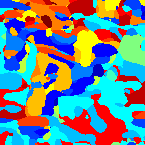}}
  \subfigure[]{\includegraphics[width=0.15\linewidth]{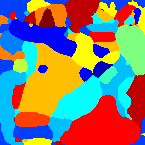}}
  \subfigure[]{\includegraphics[width=0.15\linewidth]{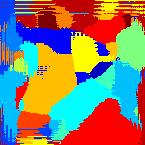}}\\
  \subfigure[]{\includegraphics[width=0.15\linewidth]{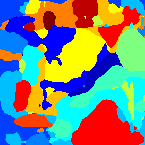}}
  \subfigure[]{\includegraphics[width=0.15\linewidth]{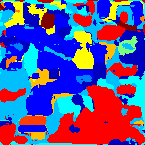}}
  \subfigure[]{\includegraphics[width=0.15\linewidth]{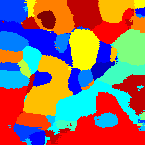}}
  \subfigure[]{\includegraphics[width=0.15\linewidth]{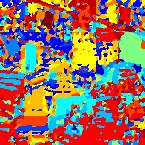}}
  \subfigure[]{\includegraphics[width=0.15\linewidth]{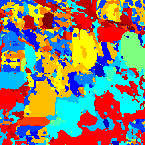}}
  \caption{Classification results of different methods on the Indian Pines dataset. (a) Original image. (b) Ground truth. (c) RBF-SVM. (d) 1-D CNN. (e) 1-D Auto-CNN. (f) 1-D P-DARTS. (g) 1-D HK-CLS. (h) 3-D CNN. (i) 3-D Auto-CNN. (j) 3-D P-DARTS. (k) NetworkAdjustment. (l) SSTN. (m) 3-D HK-CLS. (n) 3-D ANAS. (o) 3-D HK-SEG.}
  \label{acc_compare_map_indian}
\end{figure}

Since the classification maps of these datasets show similar conclusions, we only show the results of the Indian Pines dataset in Figure \ref{acc_compare_map_indian}. It is obvious that the prediction maps of 1-D pixel-level methods are strongly affected by the salt-and-pepper noise and the surfaces inside objects are too rough since only the spectral characteristics of the target pixel are considered. Nevertheless, object boundaries are retained well because the categories of edge pixels are not interfered by other pixels. Taking the consideration of spatial characteristics by incorporating more surrounding pixels, 3-D pixel-level methods produce more clean classification maps and obtain higher accuracies, but the oversmoothing phenomenon is also turning up. Compared with 3-D Auto-CNN, 3-D P-DARTS and NetworkAdjustment, our 3-D HK-CLS generates a classification map that possesses more discriminative outlines in the meantime of more pixels are correctly classified. The surface continuities inside objects of the maps generated by 3-D image-level methods are located between 1-D pixel-level methods and 3-D pixel-level methods.

 \subsubsection{Model Complexity}

 \begin{table}[t]
  \caption{Searching Cost (GPUs) and Parameter Number (K) of different networks searched on four classical datasets.}
  \newcommand{\tabincell}[2]{\begin{tabular}{@{}#1@{}}#2\end{tabular}}
  \centering
  \resizebox{0.7\linewidth}{!}{
    \begin{tabular}{|l|cc|}
    \hline
    \bfseries Indian Pines & Searching Cost  & Parameter Number  \\
    \hline
    1-D Auto-CNN & \bfseries 227.27  & \bfseries 40.67 \\
    1-D P-DARTS & 315.25 & 184.35 \\
    1-D HK-CLS& 362.74  & 3841.41 \\
    \hline
    3-D Auto-CNN& 238.46  & \bfseries 133.70 \\
    3-D P-DARTS & 586.39 & 151.70 \\
    3-D HK-CLS & \bfseries 173.14  & 249.49 \\
    \hline
    3-D ANAS & 24844.73  & 257.11 \\
    3-D HK-SEG & \bfseries 93.27  & \bfseries 109.05 \\
    \hline\hline
    \bfseries Pavia University & Searching Cost  & Parameter Number  \\
    \hline
    1-D Auto-CNN & 108.20  & \bfseries 55.32 \\
    1-D P-DARTS & 147.43 & 180.34 \\
    1-D HK-CLS& \bfseries 31.75 & 560.70 \\
    \hline
    3-D Auto-CNN& 114.41  & \bfseries 139.09 \\
    3-D P-DARTS & 352.98 & 149.83 \\
    3-D HK-CLS & \bfseries 31.68  & 161.98 \\
    \hline
    3-D ANAS & 11082.61  & 279.11 \\
    3-D HK-SEG & \bfseries 1093.78  & \bfseries 107.96 \\
    \hline\hline
    \bfseries Kennedy Space Center & Searching Cost  & Parameter Number  \\
    \hline
    1-D Auto-CNN & 301.11   & \bfseries 57.53 \\
    1-D P-DARTS & 312.30 & 183.20 \\
    1-D HK-CLS& \bfseries 94.69  & 926.38 \\
    \hline
    3-D Auto-CNN& 195.86 & 152.26 \\
    3-D P-DARTS & 554.14 & \bfseries  151.07 \\
    3-D HK-CLS & \bfseries 90.88  & 153.42 \\
    \hline
    3-D ANAS & 26428.88  & 245.34\\
    3-D HK-SEG & \bfseries 1487.67  & \bfseries 113.66\\
    \hline\hline
    \bfseries Salinas Valley & Searching Cost  & Parameter Number  \\
    \hline
    1-D Auto-CNN & 219.56  & \bfseries 65.41 \\
    1-D P-DARTS & 295.48 & 184.48 \\
    1-D HK-CLS& \bfseries 60.99  & 575.90 \\
    \hline
    3-D Auto-CNN& 214.49  & 165.08 \\
    3-D P-DARTS & 583.81 & \bfseries 151.74 \\
    3-D HK-CLS & \bfseries 99.83  & 157.27 \\
    \hline
    3-D ANAS & 24819.24  & 253.45 \\
    3-D HK-SEG & \bfseries 494.17  & \bfseries 110.59 \\
    \hline
  \end{tabular}
    }
  \label{complexity}
\end{table}

To comprehensively evaluate the proposed methods, two commonly used indices in the NAS field: searching cost and parameter number of the searched network structure are listed in Table \ref{complexity} to separately measure the time and space complexities of searching methods. Here, we count relatively competitive NAS-related comparison methods including 1-D Auto-CNN, 1-D P-DARTS, 3-D Auto-CNN, 3-D P-DARTS, and 3-D ANAS, according to their performances in Table \ref{classical_acc_compare}. Because the capacity of the 3-D image-level group is small, 3-D ANAS is also involved although it does not perform well. The unit of searching cost is GPU seconds, which are obtained by multiplying searching time and the number of used GPUs. Note in our implementations, all methods are conducted on a single GPU. Thus, the searching cost is surely the searching time. It can be seen that, except for the Indian Pines dataset, our methods have significant advantages in searching cost since the design of deriving structural parameters from hyper kernels transforms the previous complex dual optimization problem into a single one-tier optimization problem, simplifying the searching procedure and reducing searching difficulties. We can also observe that the numbers of trainable parameters in 3-D HK-SEG are significantly less than the 3-D HK-CLS since the layer number of 3-D HK-SEG is always 1 in these scenes. Compared with the 3-D ANAS adopting multiple cells for 3-D Image-level classification, with the help of unified hyper kernels that can integrate all sub kernels to represent candidate operations, and simple hierarchical modular search space, our 3-D HK-SEG is a more light network to be searched. While the architecture in the cell of 3-D Auto-CNN is complicated where many edges and nodes are contained. Therefore, the parameter numbers of the obtained 3-D HK-SEG structures are greatly smaller than 3-D ANAS. In addition, it can be observed that the values of 1-D HK-CLS are relatively large since the channel numbers are doubled when through $M/4$ blocks, where $M$ is the total block number.

  \subsubsection{Results on Challenging Scenes}

  \begin{table}[t]
    \caption{Accuracy comparisons of different methods on challenging scenes (\%)}
    \newcommand{\tabincell}[2]{\begin{tabular}{@{}#1@{}}#2\end{tabular}}
    \centering
    \resizebox{\linewidth}{!}{
    \begin{tabular}{|l|ccc|ccc|cc|}
    \hline
     Group & \multicolumn{3}{c|}{1-D Pixel-level} & \multicolumn{3}{c|}{3-D Pixel-level}  & \multicolumn{2}{c|}{3-D Image-level}  \\
     \hline\hline
    \bfseries WHU-Hi-HanChuan & \tabincell{c}{1-D \\Auto-CNN} & \tabincell{c}{1-D \\ P-DARTS} & \tabincell{c}{1-D \\ HK-CLS}  & \tabincell{c}{3-D \\ Auto-CNN} & \tabincell{c}{3-D \\ P-DARTS}  & \tabincell{c}{3-D \\ HK-CLS} & \tabincell{c}{3-D\\ANAS} & \tabincell{c}{3-D\\HK-SEG}  \\
    \hline
    OA  & \tabincell{c}{57.86\\$\pm$2.89} & \tabincell{c}{57.18\\$\pm$2.32} & \bfseries\tabincell{c}{58.44\\$\pm$1.85} & \tabincell{c}{73.34\\$\pm$5.75} & \tabincell{c}{79.60\\$\pm$3.08}  & \bfseries\tabincell{c}{80.48\\$\pm$2.06} & \tabincell{c}{62.04\\$\pm$2.39}  &  \bfseries\tabincell{c}{67.23\\$\pm$3.26} \\
    AA  & \bfseries\tabincell{c}{54.81\\$\pm$2.47}  & \tabincell{c}{53.25\\$\pm$1.58} & \tabincell{c}{52.72\\$\pm$1.39} & \tabincell{c}{72.14\\$\pm$3.44}  & \tabincell{c}{77.92\\$\pm$2.14}  & \bfseries\tabincell{c}{78.78\\$\pm$1.77} & \tabincell{c}{53.81\\$\pm$2.34}  & \bfseries\tabincell{c}{62.13\\$\pm$2.78}  \\
    Kappa & \tabincell{c}{52.27\\$\pm$3.09} & \tabincell{c}{51.56\\$\pm$2.39} & \bfseries\tabincell{c}{52.80\\$\pm$1.94} & \tabincell{c}{69.46\\$\pm$6.29}  & \tabincell{c}{76.44\\$\pm$3.40}  & \bfseries\tabincell{c}{77.50\\$\pm$2.24} & \tabincell{c}{56.29\\$\pm$2.55} & \bfseries\tabincell{c}{62.38\\$\pm$3.54}  \\
    \hline\hline
    \bfseries WHU-Hi-HongHu & \tabincell{c}{1-D \\Auto-CNN} & \tabincell{c}{1-D \\ P-DARTS} & \tabincell{c}{1-D \\ HK-CLS}  & \tabincell{c}{3-D \\ Auto-CNN} & \tabincell{c}{3-D \\ P-DARTS} & \tabincell{c}{3-D \\ HK-CLS} & \tabincell{c}{3-D\\ANAS} & \tabincell{c}{3-D\\HK-SEG}  \\
    \hline
    OA  & \tabincell{c}{54.16\\$\pm$4.75} & \tabincell{c}{55.66\\$\pm$3.82} & \bfseries\tabincell{c}{57.01\\$\pm$2.77} & \tabincell{c}{80.46\\$\pm$1.26}  & \tabincell{c}{85.81\\$\pm$1.38}  & \bfseries\tabincell{c}{87.32\\$\pm$1.53} & \tabincell{c}{64.61\\$\pm$3.58}   & \bfseries\tabincell{c}{73.10\\$\pm$3.41}  \\
    AA  & \bfseries\tabincell{c}{53.34\\$\pm$1.70} & \tabincell{c}{53.02\\$\pm$1.72} & \tabincell{c}{51.83\\$\pm$1.61} & \tabincell{c}{77.02\\$\pm$2.62} & \tabincell{c}{82.36\\$\pm$2.32}  & \bfseries\tabincell{c}{85.59\\$\pm$1.54} & \tabincell{c}{64.52\\$\pm$2.76}  & \bfseries\tabincell{c}{69.94\\$\pm$4.31}  \\
    Kappa & \tabincell{c}{47.33\\$\pm$4.37} & \tabincell{c}{48.60\\$\pm$3.58} & \bfseries\tabincell{c}{51.26\\$\pm$2.86} & \tabincell{c}{75.81\\$\pm$1.45}  & \tabincell{c}{82.24\\$\pm$1.71}  & \bfseries\tabincell{c}{84.19\\$\pm$1.83} & \tabincell{c}{58.34\\$\pm$3.88} &  \bfseries\tabincell{c}{67.10\\$\pm$4.46} \\
    \hline
  \end{tabular}
      }
    \label{challenge_acc_compare}
  \end{table}

  \begin{figure}[t]
    \centering
    \subfigure[]{\includegraphics[width=0.12\linewidth]{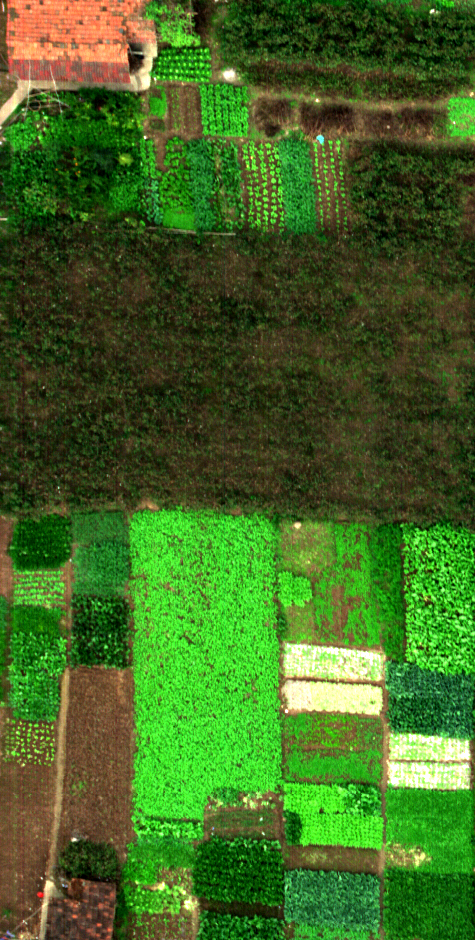}}
    \subfigure[]{\includegraphics[width=0.12\linewidth]{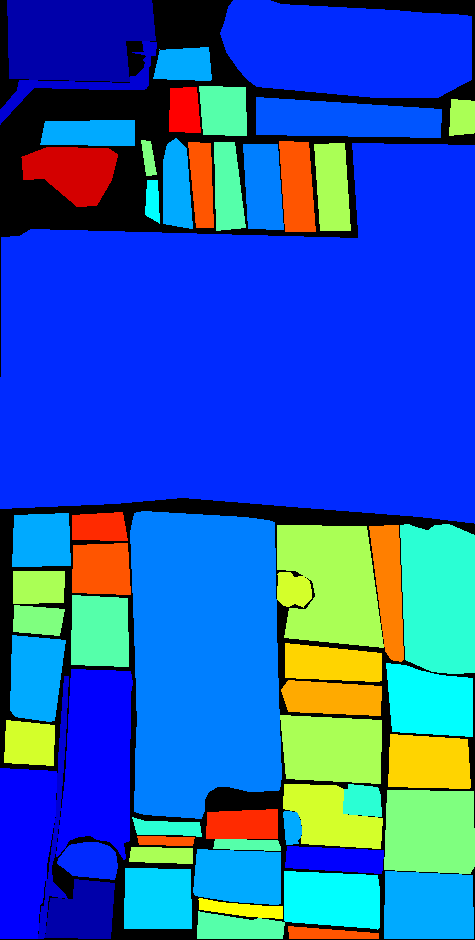}}
    \subfigure[]{\includegraphics[width=0.12\linewidth]{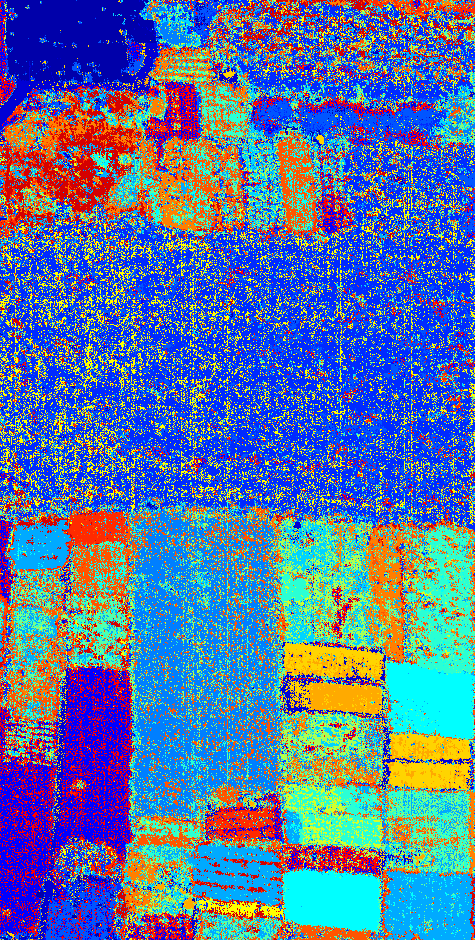}}
    \subfigure[]{\includegraphics[width=0.12\linewidth]{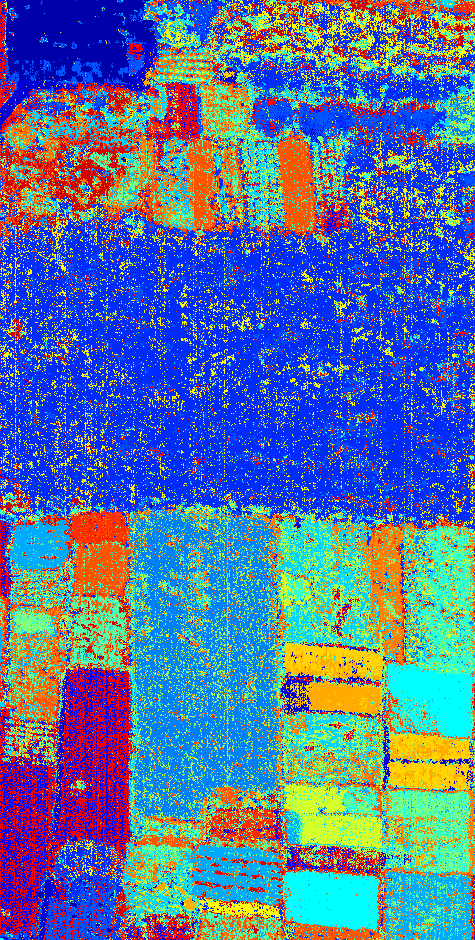}}
    \subfigure[]{\includegraphics[width=0.12\linewidth]{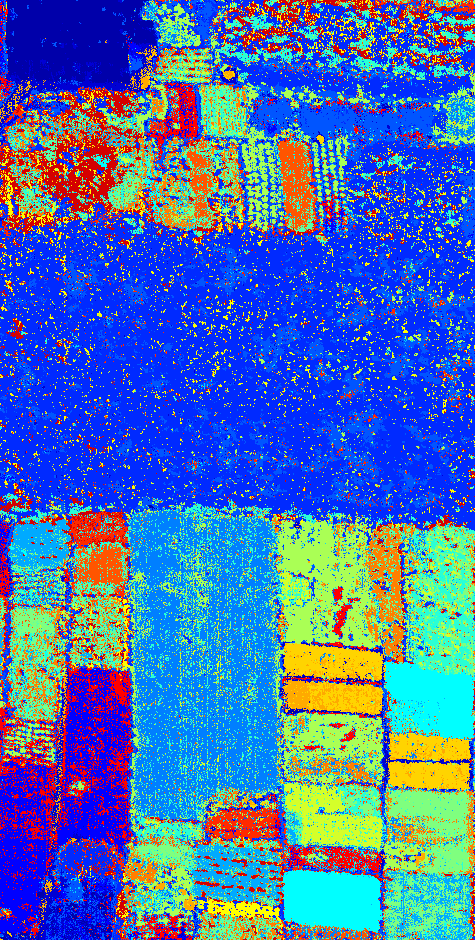}}\\
    \subfigure[]{\includegraphics[width=0.12\linewidth]{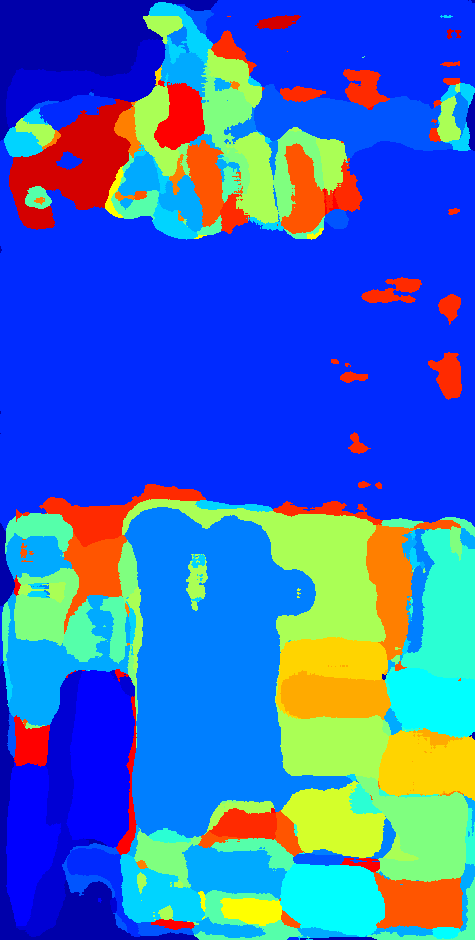}}
    \subfigure[]{\includegraphics[width=0.12\linewidth]{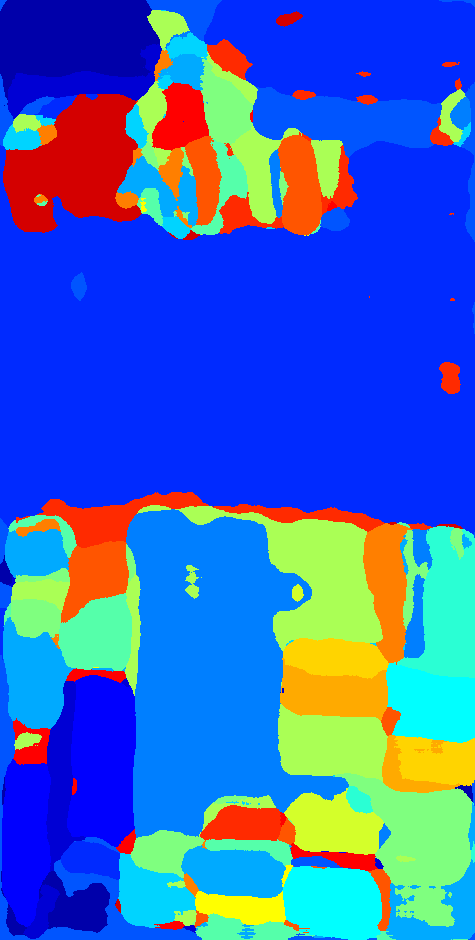}}
    \subfigure[]{\includegraphics[width=0.12\linewidth]{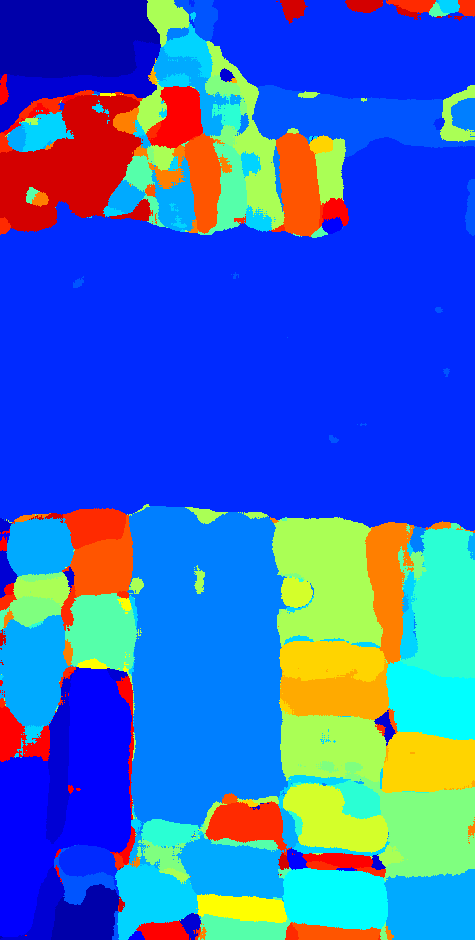}}
    \subfigure[]{\includegraphics[width=0.12\linewidth]{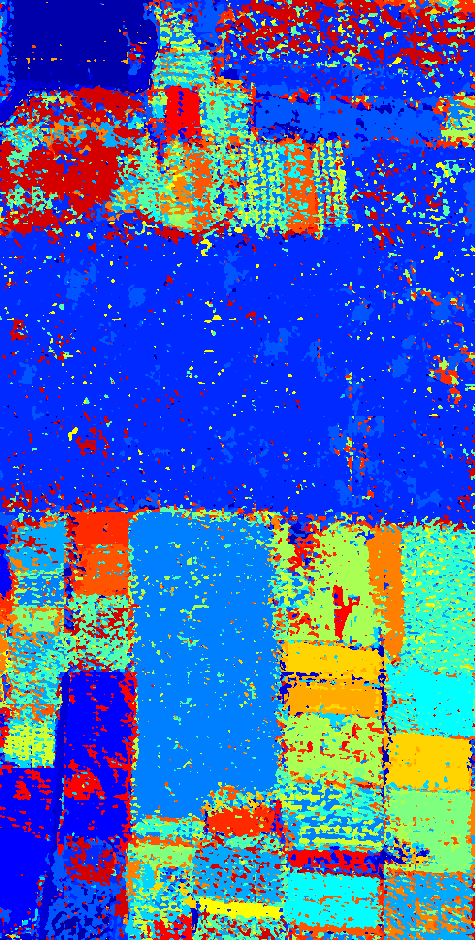}}
    \subfigure[]{\includegraphics[width=0.12\linewidth]{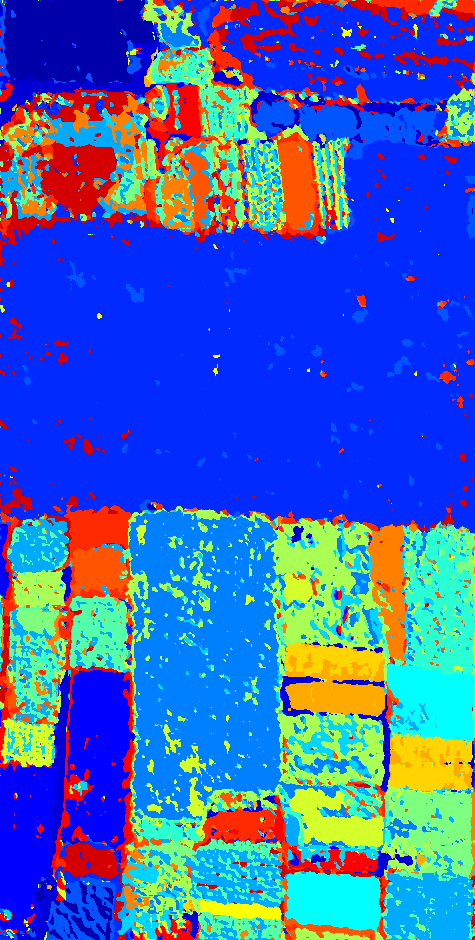}}
    \caption{Classification results of different methods on the WHU-Hi-HongHu dataset. (a) Original image. (b) Ground truth. (c) 1-D Auto-CNN. (d) 1-D P-DARTS. (e) 1-D HK-CLS. (f) 3-D Auto-CNN. (g) 3-D P-DARTS. (h) 3-D HK-CLS. (i) 3-D ANAS. (j) 3-D HK-SEG.}
    \label{acc_compare_map_honghu}
  \end{figure}

  Besides classical HSI classification datasets, we further evaluate the proposed methods and compare them with the aforementioned competitive methods on two more complex scenes: HanChuan and HongHu of the WHU-Hi dataset. At this time, the block number and layer number $(M, N)$ of 1-D HK-CLS on HanChuan and HongHu scenes are separate as $(3,3)$ and $(3,1)$. Similarly, $(3,2)$ and $(3,3)$ for 3-D HK-CLS, while 3-D HK-SEG adopts $(3,1)$. The 3-D convolution forms of 3-D HK-CLS and 3-D HK-SEG are ``1-D Conv / 2-D DW Conv'' and ``3-D Conv'', which are not changed by scenes. The determinations of hyperparameters and 3-D convolution forms are similar to the experiments in Section IV-C and D. Table \ref{challenge_acc_compare} shows the classification accuracy, from which we can observe that even if on more challenging scenes, the proposed methods can still perform well and produce discriminative classification maps, as shown in Figure \ref{acc_compare_map_honghu}.

 \subsection{Understanding Searching and Training Procedures}

 \begin{figure}[t]
  \centering
  \subfigure[]{\includegraphics[width=0.42\linewidth]{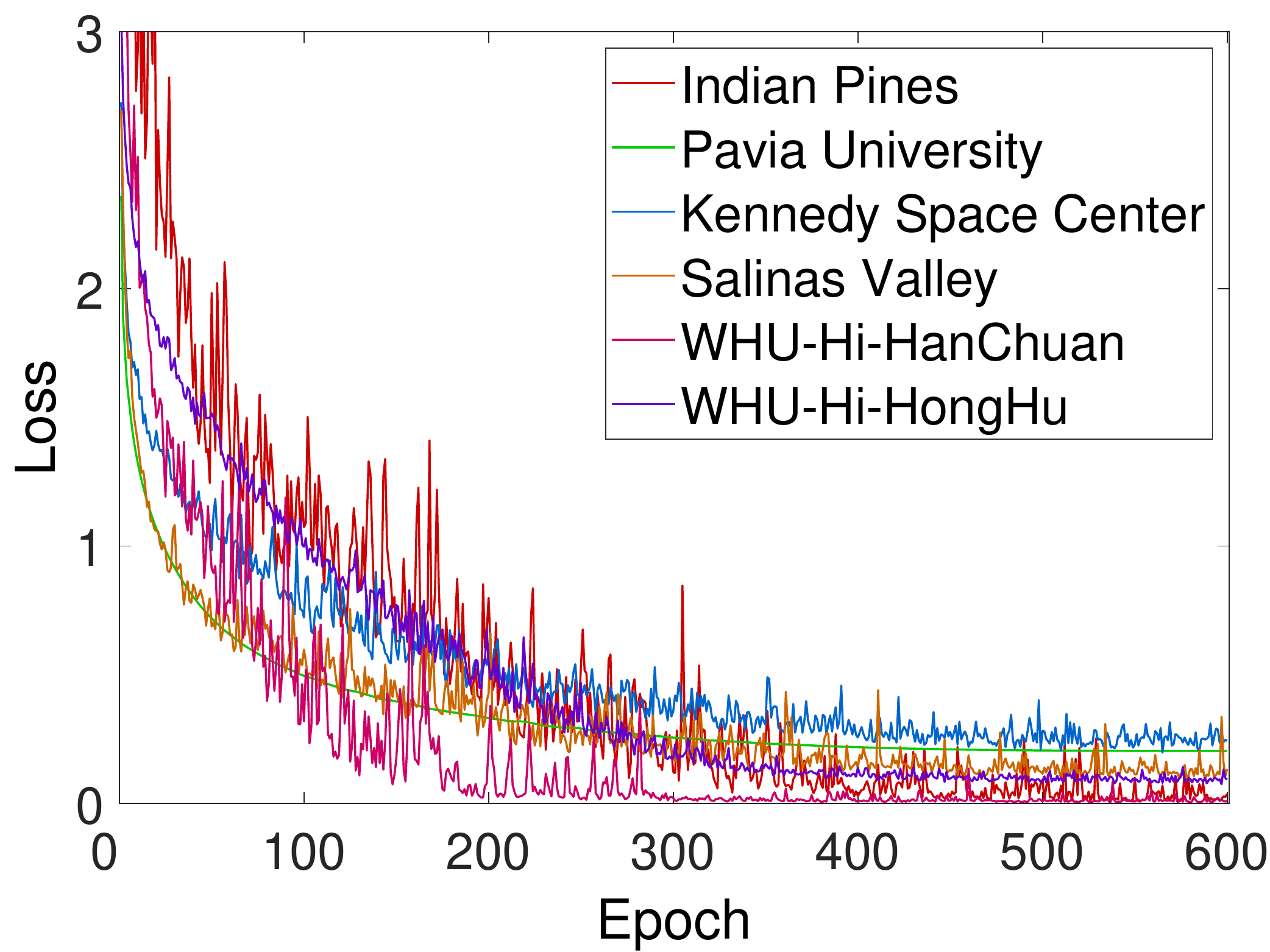}}
  \subfigure[]{\includegraphics[width=0.42\linewidth]{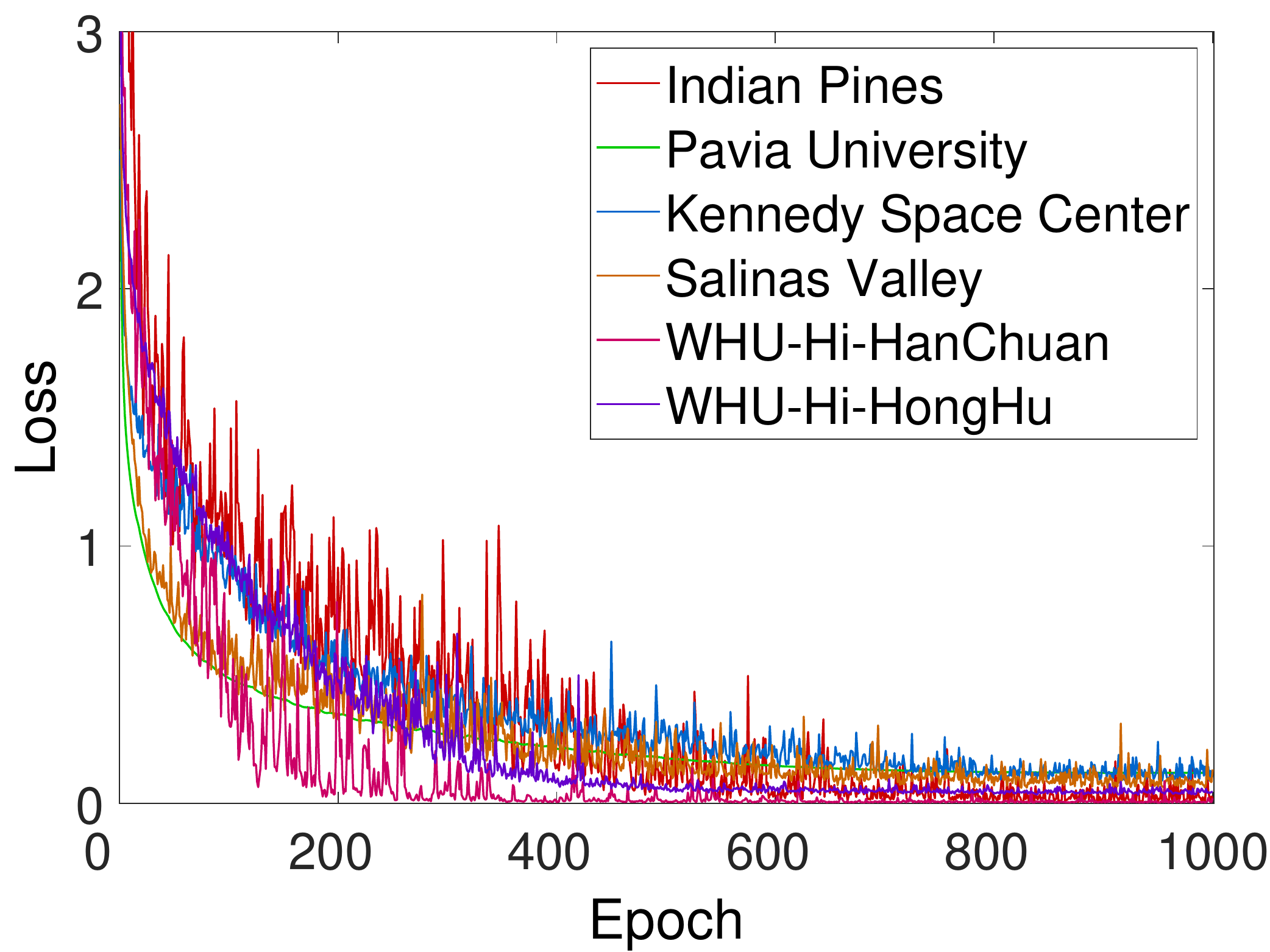}}\\
  \subfigure[]{\includegraphics[width=0.42\linewidth]{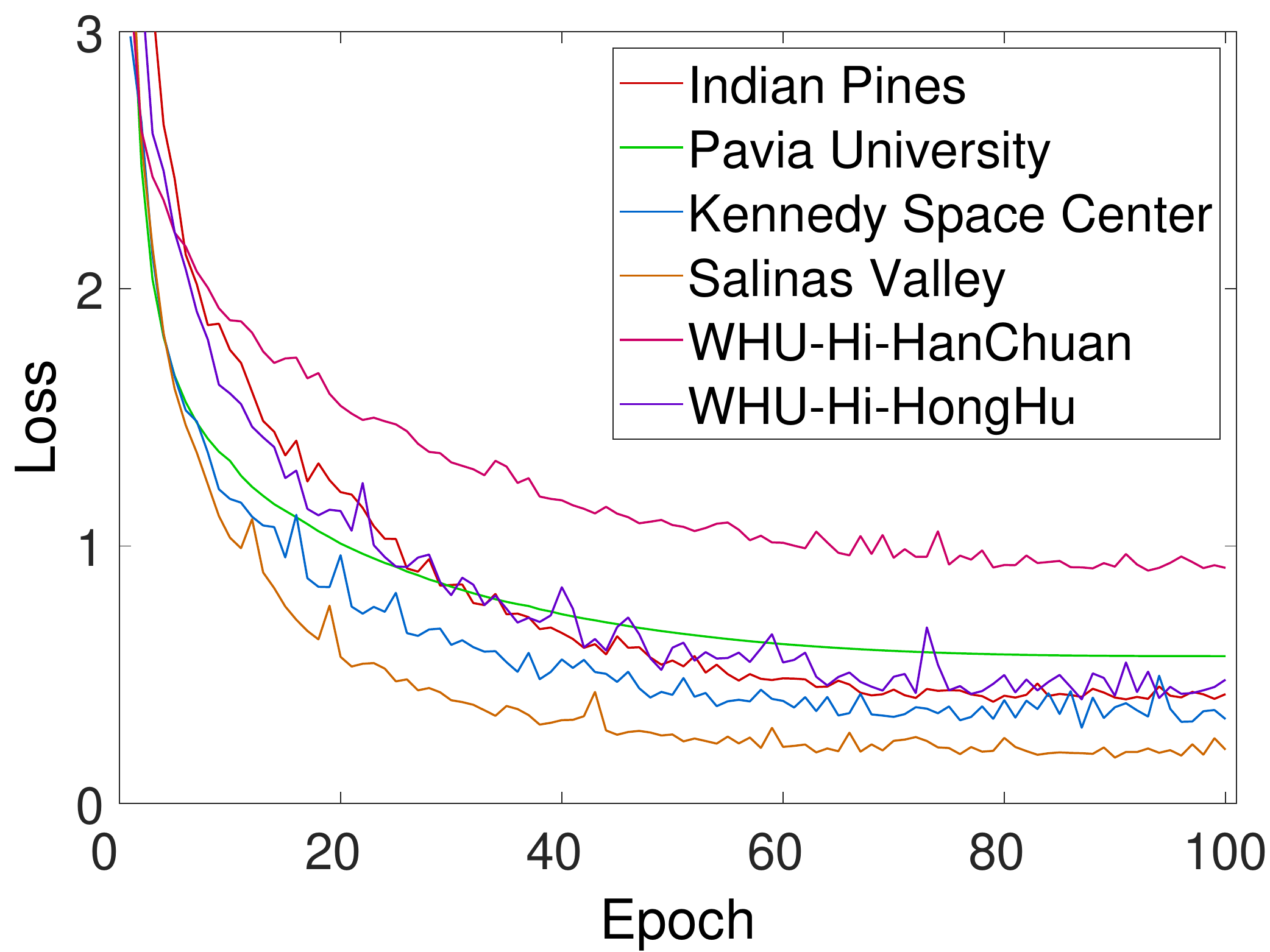}}
  \subfigure[]{\includegraphics[width=0.42\linewidth]{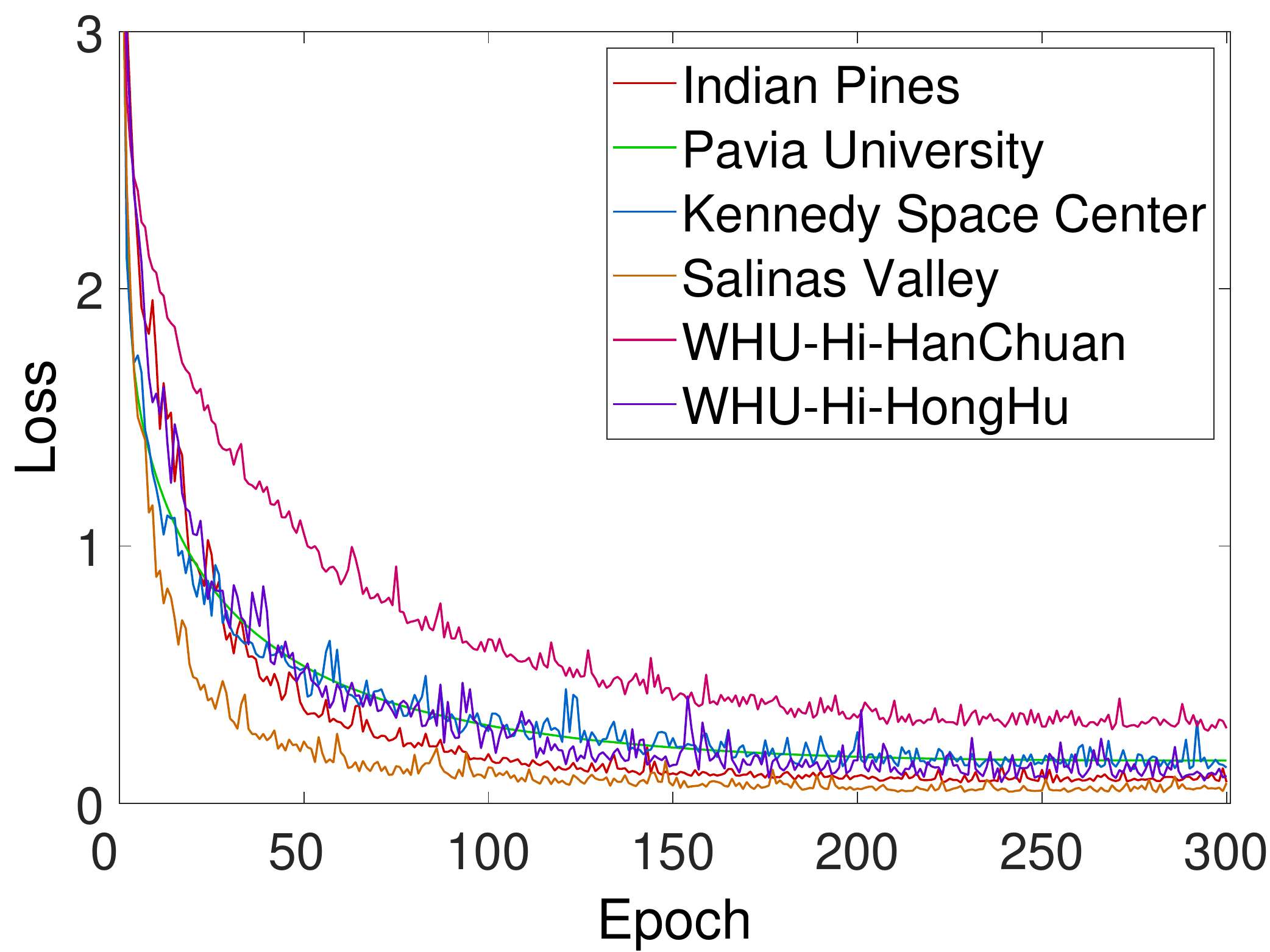}}\\
  \subfigure[]{\includegraphics[width=0.42\linewidth]{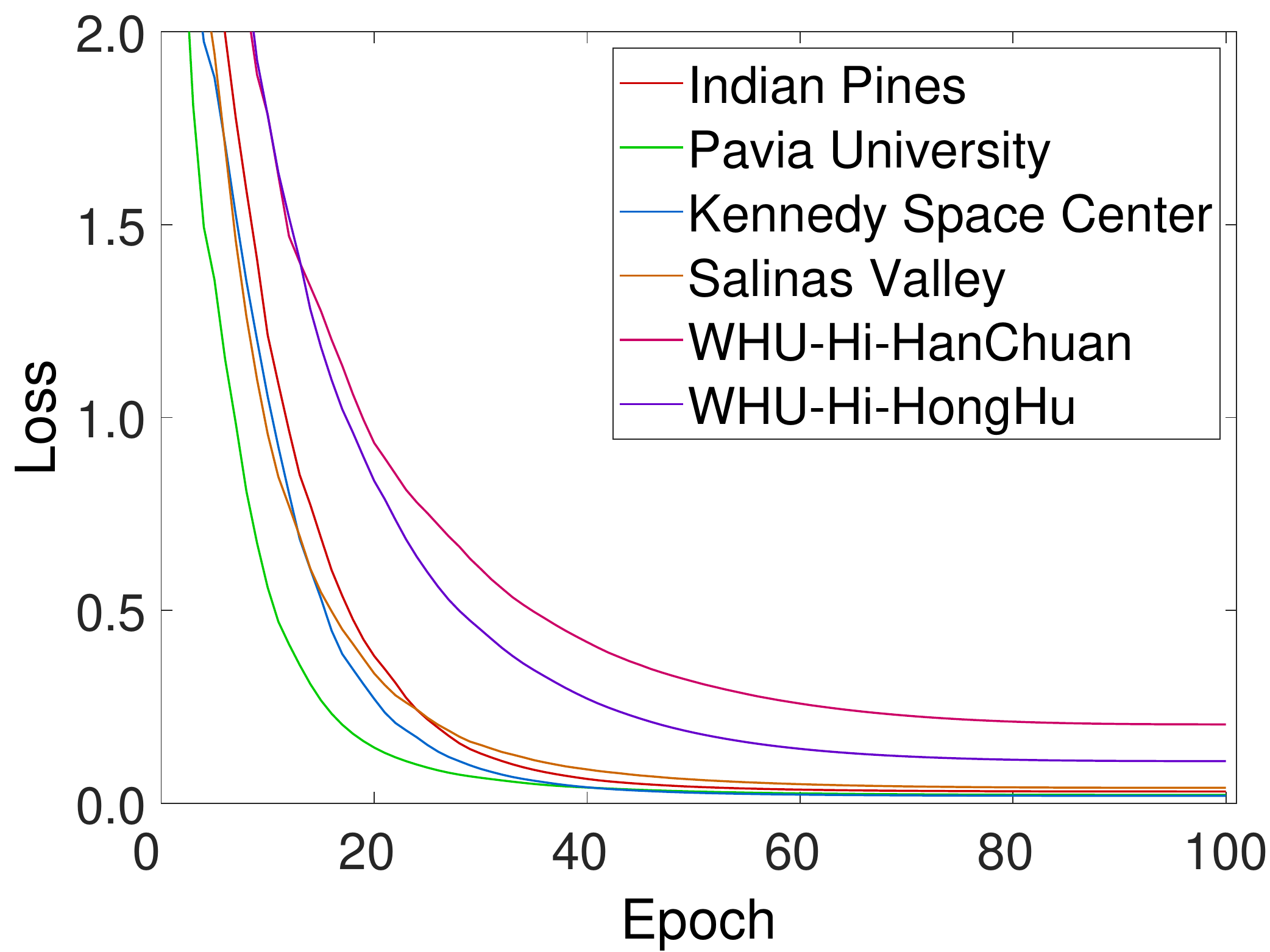}}
  \subfigure[]{\includegraphics[width=0.42\linewidth]{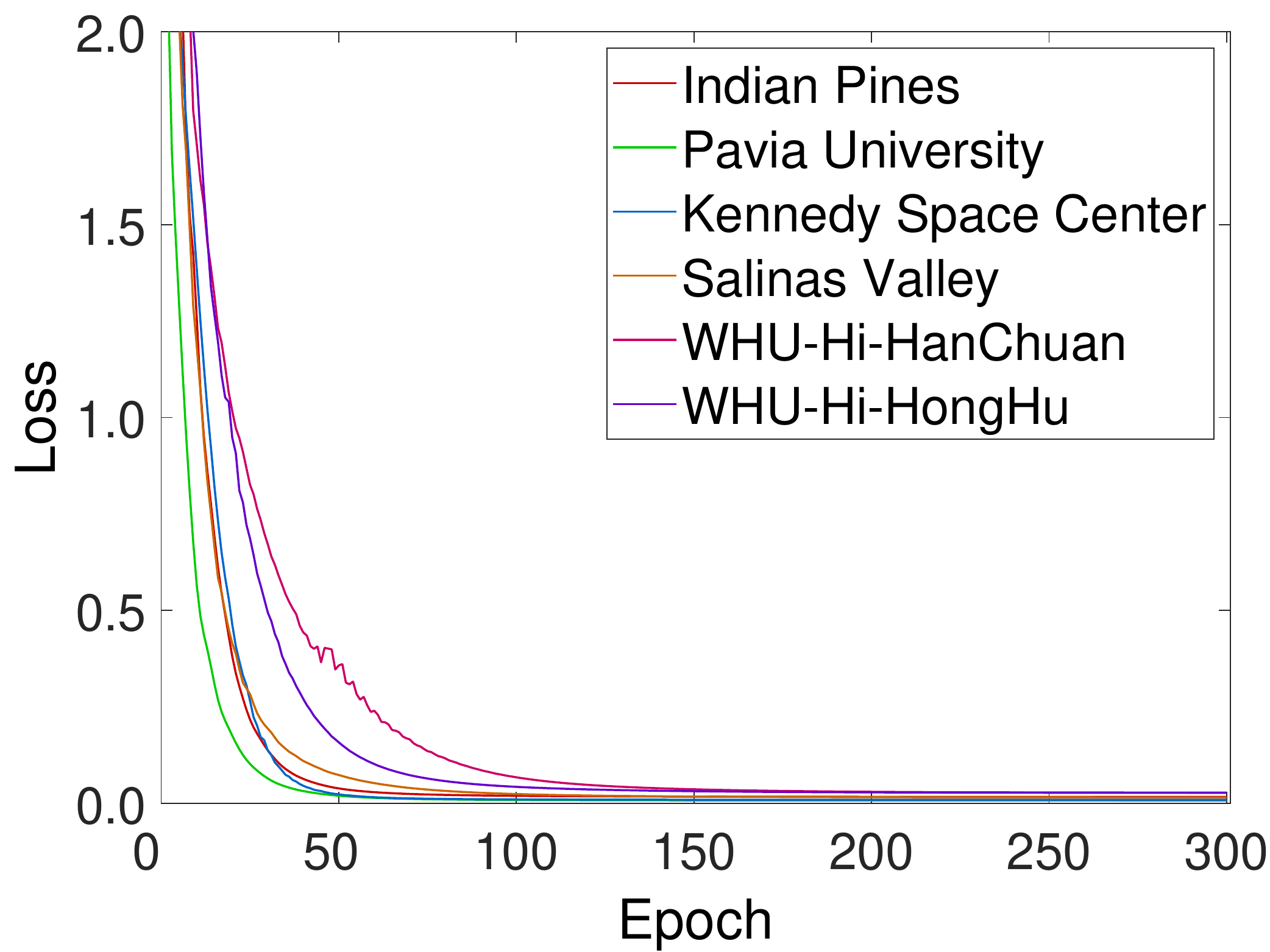}}
  \caption{The loss curves in searching and training procedures of the proposed networks on different datasets: (a)-(b) 1-D HK-CLS. (c)-(d) 3-D HK-CLS. (e)-(f) 3-D HK-SEG. (a)(c)(e) are the searching procedures, while the training procedures are presented in (b)(d)(f).}
  \label{loss}
\end{figure}

To better understand the proposed methods, we monitor the loss declining of the above three kinds of networks in searching and training procedures, and the losses are plotted in Figure \ref{loss}. It can be seen that all networks have reached convergence. Specifically, the loss profiles on the Pavia University dataset are very smooth, indicating the networks have achieved fantastic fitting on training set, while the HanChuan scene is the opposite, showing the complexity of this dataset. However, Table \ref{classical_acc_compare} shows the accuracies of all methods including ours on the Pavia University scene are not very well because this dataset has the least training samples, causing the overfitting. The smoother loss curves and worse performances of 3-D HK-SEG compared with 3-D HK-CLS also demonstrate this point. In our HKNAS, since only the weights of hyper kernels need to be optimized, causing the optimizations in searching procedures are similar to training procedures, where only network weights are updated. Therefore, the shapes of loss curves in searching and training stages are in similar patterns.

\subsection{Visualization of Searched Architecture}

At last, we visualize the obtained architectures in Table \ref{arch}, where the row and column number of the matrices separately represents the number of blocks and the number of layers inside each block. For a matrix $A$, the value of $A_{ij}$ is the index of the selected operation in the HKNAS part at the $i$th block and the $j$th layer. For 1-D HK-CLS, $A_{ij} = s^*-1 $, where $s^*$ is the index of sub kernels and structural parameters in formula (\ref{index_12d}) and represents the $1 \times (2s^*+1)$ convolution. For 3-D HK-CLS and 3-D HK-SEG, if standard 3-D convolutions are adopted, $A_{ij}$ means a 3-D convolution whose kernel size is $\left(2(A_{ij}+1)+1\right) \times (2(A_{ij}+1)+1) \times (2(A_{ij}+1)+1)$. Since ``/'' is the parallel form, $a$/$b$ represents the 3-D convolution is substituted by two parallel 1-D convolution and 2-D depth-wise convolutional branches, and $a,b$ are their indices, respectively. The serial decomposition of a 3-D convolution is symbolized in a form of $10 \times a + b$, where $a$ and $b$ are the indices of two candidate 1-D convolution or 2-D depth-wise convolution operations before and after, respectively. Therefore, if a network adopts serial 3-D convolution forms, $a$ and $b$ are $\left\lfloor A_{ij}/10 \right\rfloor$ and $\bmod{(A_{ij},10)}$, respectively.

\begin{table}[t]
  \caption{The final obtained architectures searched on the used datasets with the proposed methods}
  \newcommand{\tabincell}[2]{\begin{tabular}{@{}#1@{}}#2\end{tabular}}
  \centering
  \resizebox{0.8\linewidth}{!}{
  \begin{tabular}{|l|c|c|c|}
  \hline
  Architecture &  1-D HK-CLS & 3-D HK-CLS & 3-D HK-SEG\\
  \hline
  Indian Pines & $\left[\begin{array}{c c c c c}
    1 & 1 & 3& 2& 3 \\
    3 & 1 & 0 & 1& 1\\
    1 & 2 & 0 & 3 & 1\\
    3 &1 &1 &0 & 1  \\
    2 & 2 & 3 & 0 & 0\\
    3 &0 &2 &2 &2 \\
   \end{array} \right]$ & $\left[\begin{array}{c c c c}
    2 & 1 & 2& 1 \\
    0 & 0& 2 & 0\\
    0 & 2 & 2 & 1\\
   \end{array} \right]$ & $\left[\begin{array}{c}
    1 \\
    0 \\
    2 \\
   \end{array} \right]$ \\
   \hline
  Pavia University & $\left[\begin{array}{c}
    0 \\
    3 \\
    2 \\
    2  \\
   \end{array} \right]$ & $\left[\begin{array}{cc}
    3/3 & 0/1 \\
    0/0 & 2/1 \\
    2/0 & 0/2 \\
   \end{array} \right]$ & $\left[\begin{array}{c}
    23 \\
    23 \\
    23 \\
   \end{array} \right]$ \\
   \hline
  Kennedy Space Center & $\left[\begin{array}{cc}
    3 & 1 \\
    2 & 0 \\
    2 & 1 \\
   \end{array} \right]$ & $\left[\begin{array}{cc}
    0 & 1 \\
    3 & 0 \\
    0 & 3 \\
   \end{array} \right]$ & $\left[\begin{array}{c}
    30 \\
    12 \\
    30 \\
   \end{array} \right]$ \\
   \hline
  Salinas Valley & $\left[\begin{array}{c}
    1 \\
    1 \\
    0 \\
    2  \\
   \end{array} \right]$ & $\left[\begin{array}{cc}
    0 & 2 \\
    0 & 3 \\
    2 & 1 \\
   \end{array} \right]$ & $\left[\begin{array}{c}
    0\\
    3 \\
    1 \\
   \end{array} \right]$ \\
   \hline
  WHU-Hi-HanChuan &
  $\left[\begin{array}{c c c}
    2 & 0 & 0 \\
    1 & 1 & 3\\
    3 & 1 & 2\\
   \end{array} \right]$ & $\left[\begin{array}{cc}
    3/2 & 0/3 \\
    1/0 & 0/2 \\
    0/1 & 3/1 \\
   \end{array} \right]$ & $\left[\begin{array}{c}
    0 \\
    1 \\
    0 \\
   \end{array} \right]$ \\
   \hline
  WHU-Hi-HongHu &
  $\left[\begin{array}{c}
    0 \\
    0 \\
    2 \\
   \end{array} \right]$ &$\left[\begin{array}{ccc}
    1/0 & 1/1 & 3/2 \\
    1/3 & 2/0 & 2/3 \\
    3/2 & 3/2 & 1/3 \\
   \end{array} \right]$ & $\left[\begin{array}{c}
    0 \\
    0 \\
    0 \\
   \end{array} \right]$\\
  \hline
\end{tabular}
  }
  \label{arch}
\end{table}

\section{Conclusion}

In this paper, three types of networks whose structures are automatically searched to effectively classify the HSI from pixel-level or image-level using 1-D or 3-D convolutions, respectively. Concretely, we first argue that the areas with different shapes and sizes of existing super kernels can be regarded as sub kernels, and elaborate the implicit connections between architectures and operators in NAS procedures. Then, we propose the hyper kernel which can further derive structural parameters to search for suitable architectures compared with super kernels, so as to convert the previous complex dual optimization problem to a simple one-tier searching task, where only the weights of hyper kernels need to be optimized. To this end, we imagine the simplest case, that is, there are only multiple standard convolutions of different kernel sizes centered on the target pixel, and we consider them as candidate operations. Then, the concepts of mask matrix and core area are introduced to generate structural parameters. Besides producing the structural parameters of 1-D and 2-D candidate convolution operations, we also discuss the situation of 3-D convolutions and propose to combine 3-D convolution decomposition technologies with the presented hyper kernel search strategy to obtain abundant network structures while the performances of 3-D convolutions are also maintained. To better implement the proposed hyper kernel scheme, we design a hierarchical multi-module search space where only convolutions are involved to fully integrate candidate operations into hyper kernels. By simultaneously employing the proposed searching strategy and searching space, diverse architectures are finally obtained to flexibly and efficiently realize HSI classifications.

To evaluate the proposed methods, extensive experiments on six commonly used HSI classification datasets are implemented. We first determine optimal macro network structures including the numbers of blocks or layers and microarchitectures in the HKNAS part such as the 3-D convolution forms of each scene by conducting hyperparameter analyses and ablation studies. Then the effectiveness of the obtained networks compared with existing advanced methods is proved through comprehensive accuracy and complexity evaluations. The analyses of model status monitoring in searching and training procedures and the visualizations of the final obtained architectures further demonstrate the persuasiveness of the proposed methods.

\ifCLASSOPTIONcaptionsoff
  \newpage
\fi

\bibliographystyle{IEEEtran}
% argument is your BibTeX string definitions and bibliography database(s)
\bibliography{HSI_HKNAS}

\end{document}